\documentclass[11pt]{article}

\usepackage[final]{acl}

\usepackage{times}
\usepackage{latexsym}

\usepackage[T1]{fontenc}

\usepackage[utf8]{inputenc}

\usepackage{microtype}

\usepackage{inconsolata}

\usepackage{graphicx}
\usepackage{subcaption}

\usepackage{soul}
\usepackage{mwe}
\usepackage{amsmath}
\usepackage{amssymb}
\usepackage{tabularx}
\usepackage{arydshln}
\usepackage{booktabs}
\usepackage{adjustbox}
\usepackage{algorithm}
\usepackage{algpseudocode}
\usepackage{standalone}
\usepackage{tikz}
\usetikzlibrary{arrows.meta, positioning, calc, bending}

\newif\ifshowdiscussion

\showdiscussionfalse

\newcommand{\dav}[1]{\ifshowdiscussion {\color{blue} \textbf{DAV:} \textit{#1} }\fi}

\newcommand{\felixcomment}[1]{\ifshowdiscussion {\color{purple} \textbf{FE:} \textit{#1}} \fi}
\newcommand{\felix}[1]{{\ifshowdiscussion \color{purple}\fi#1}}
\newcommand{\felixremove}[1]{\ifshowdiscussion {\color{purple!20}#1} \else \fi}

\title{Online Language Adaptive Sampling for \\Better Distributed Cross-lingual Gains}

\author{Quang Phuoc Nguyen$^1$\thanks{Equal contribution.}, Felix Gaschi$^{2*}$, David Anugraha$^{3}$, \\
\textbf{Santiago Martínez Novoa}$^{4}$, \textbf{En-Shiun Annie Lee}$^{1, 5}$ \\
  $^{1}$Ontario Tech University$\quad^{2}$Doctrine$\quad^{3}$Stanford University\\ $\quad^{4}$University of the Andes$\quad^{5}$University of Toronto\\
\texttt{quangphuoc.nguyen@ontariotechu.net, felix.gaschi@doctrine.fr,}\\ \texttt{david.anugraha@stanford.edu, s.martinezn@uniandes.edu.co,}\\ \texttt{Annie.Lee@ontariotechu.ca}}

\begin{document}
\maketitle
\begin{abstract}

Realignment is a promising approach for improving the cross-lingual transfer ability of multilingual language models, particularly for extremely low-resource languages (LRLs). However, existing realignment methods rely on uniform and random sampling of parallel sentences across languages, which may be suboptimal under limited batch sizes. In practice, models may benefit from seeing certain languages more frequently, especially those that are poorly aligned, and the optimal distribution can evolve throughout training. In this work, we propose a simple yet effective adaptive sampling strategy that assigns trainable sampling probabilities to each language. Languages that contribute more to the realignment loss are sampled more frequently in subsequent batches, and the optimal distribution can evolve throughout training. Our method employs an inner–outer optimization loop with a small overhead, leading to consistent performance improvements and, more importantly, distributing the gains across languages. We observed a $+0.67$ average performance increase on all tasks with XLM-R, and $+0.60$ with Gemma 2 9B compared with uniform realignment. Furthermore, our method is robust across different models.\footnote{Code available at \url{https://github.com/felixgaschi/multilingual-alignment-and-transfer}}

\end{abstract}

\section{Introduction}

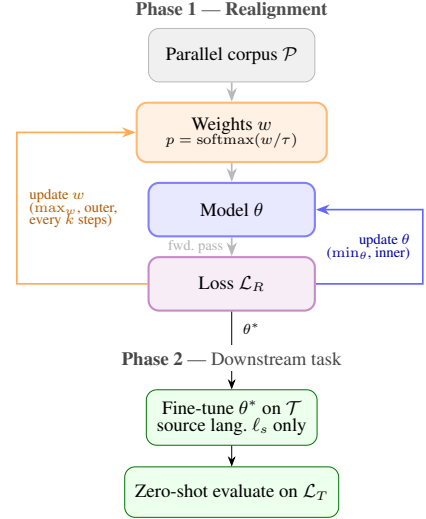
\begin{figure}[t]
    \centering
    \begin{tikzpicture}[
    >=Stealth,
    box/.style = {
        draw, rounded corners = 5pt,
        minimum width = 2.8cm, minimum height = 0.9cm,
        align = center, font = \small, inner sep = 5pt
    },
    data/.style   = {box, fill=gray!12,  draw=gray!55},
    mbox/.style   = {box, fill=blue!9,   draw=blue!50,  line width=1pt},
    lbox/.style   = {box, fill=violet!7, draw=violet!45, line width=1pt},
    wpbox/.style  = {box, fill=orange!11, draw=orange!55, line width=1pt,
                     minimum width=3.2cm, minimum height=1.0cm},
    tbox/.style   = {box, fill=green!7,  draw=green!45!black,
                     minimum width=2.8cm, minimum height=0.85cm},
    fwd/.style      = {->, thick, gray!55},
    innerupd/.style = {->, thick, blue!60},
    outerupd/.style = {->, thick, orange!65},
]

\node[font=\small\bfseries, gray!60!black] at (0, 0.75)
    {Phase 1 \normalfont---\bfseries{} Realignment};

\node[data]  (data)  at (0,  0.0)  {Parallel corpus $\mathcal{P}$};
\node[wpbox] (wp)    at (0, -1.3)
    {Weights $w$\\[-3pt]
     \scriptsize $p = \operatorname{softmax}(w/\tau)$};
\node[mbox]  (model) at (0, -2.6)  {Model $\theta$};
\node[lbox]  (loss)  at (0, -3.85) {Loss $\mathcal{L}_R$};

\draw[fwd] (data.south)  -- (wp.north);
\draw[fwd] (wp.south)    -- (model.north);
\draw[fwd] (model.south) --
    node[left, font=\scriptsize, gray!60] {fwd.\ pass}
    (loss.north);

\draw[innerupd] (loss.east) -- ++(1.8,0) -- ++(0,1.25) -- (model.east);
\node[left=0.08cm, font=\scriptsize, blue!65!black, align=right]
    at ($(loss.east)+(1.8, 0.625)$)
    {update $\theta$\\[-2pt]($\min_\theta$, inner)};

\draw[outerupd] (loss.west) -- ++(-2.2,0) -- ++(0,2.55) -- (wp.west);
\node[right=0.08cm, font=\scriptsize, orange!70!black, align=left]
    at ($(loss.west)+(-2.2, 1.275)$)
    {update $w$\\[-2pt]($\max_w$, outer,\\[-2pt]every $k$ steps)};

\node[tbox] (finetune) at (0, -6.1)
    {Fine-tune $\theta^*$ on $\mathcal{T}$\\[-2pt]
     \footnotesize source lang.\ $\ell_s$ only};

\node[tbox, below=0.35cm of finetune] (eval)
    {Zero-shot evaluate on $\mathcal{L}_T$};

\draw[->] (loss.south) -- (finetune.north);
\node[right=0.05cm, font=\scriptsize]
    at ($(loss.south)!0.22!(finetune.north)$) {$\theta^*$};
\node[fill=white, draw=none, inner sep=3pt,
      font=\small\bfseries, text=gray!55!black]
    at ($(loss.south)!0.60!(finetune.north)$)
    {Phase 2 \normalfont{---} Downstream task};

\draw[->] (finetune) -- (eval);

\end{tikzpicture}
    \caption{Overview of our method. Phase 1 depicts the realignment process, where the model is trained on parallel data using a contrastive loss. Phase 2 illustrates the downstream cross-lingual transfer.}
    \label{fig:method}
\end{figure}

Multilingual language models (MLMs) exhibit strong zero-shot cross-lingual transfer, yet still struggle on linguistically distant low-resource languages (LRLs)~\citep{pires-etal-2019-multilingual}. Prior studies attribute much of this failure to weak cross-lingual representation alignment: when language representations are poorly aligned, downstream performance drops sharply~\citep{gaschi-etal-2023-exploring, kargaran-etal-2025-mexa}. Contrastive realignment on parallel data has emerged as a data-efficient mitigation~\citep{cao2020multilingualalignmentcontextualword, wu-dredze-2020-explicit}, attractive given the limited data available for LRLs~\citep{anugraha-etal-2025-proxylm}. Yet its benefits remain inconsistent: some languages improve substantially while others see little gain or even regression \citep{wu-dredze-2020-explicit, Efimov_2023}.

Prior work on improving realignment has focused mostly on the algorithmic side~\citep{gaschi-etal-2023-exploring, bakos-etal-2025-alignfreeze}, with little attention to efficient data utilization. Yet strategic data selection improves generalization, efficiency, and robustness across many settings \citep{wang2019target, xie2023doremioptimizingdatamixtures, albalak2023efficientonlinedatamixing, chen-etal-2025-scale, anugraha2025r3}. The closest work, \citet{nguyen-etal-2025-rethinking}, found that the choice of included languages matters, but their selection is static and fixed before training, ignoring that realignment needs may shift as training progresses.

In this work, we propose a dynamic language-wise sampling strategy for the realignment phase. As shown in Figure~\ref{fig:method}, rather than sampling languages uniformly, we learn per-language weights from the realignment loss, adaptively oversampling languages that benefit most from alignment (Phase 1), then evaluate via standard zero-shot cross-lingual transfer (Phase 2). Our results show consistent improvements over uniform and static baselines, with gains increasing at scale and distributing more evenly across languages. We further show that the approach generalizes to decoder-only models, extending dynamic sampling beyond encoder-only classification settings.

\section{Methodology}

In our setup, only the source language (English) carries task supervision; we
evaluate zero-shot transfer to a set of target languages $\mathcal{L}_T$. Since
target-language labels are unavailable, we use a contrastive realignment loss
over parallel data as a language-agnostic proxy that pulls cross-lingual
sentence representations together, following \citet{wu-dredze-2020-explicit}.

Following \citet{wu-dredze-2020-explicit, gaschi-etal-2023-exploring}, for a
batch $\mathcal{P}$ of parallel pairs $(x, y)$ of source and target
representations,

\begin{equation}\label{eq:realignment}
L_R = -\frac{1}{2|\mathcal{P}|}\sum_{(x,y)\in\mathcal{P}}\bigl[\log\sigma(x,y) + \log\sigma(y,x)\bigr]
\end{equation}

where $\sigma(x,y)$ is the in-batch contrastive softmax over cosine similarities
with temperature $T_R=0.1$ (full form in
Appendix~\ref{appendix:realignment-loss}). The two terms enforce alignment
symmetrically.

\subsection{Adaptive Language Sampling}

\begin{algorithm}[t]
\caption{Adaptive Sampling for Language-Weighted Realignment}
\label{alg:bilevel}
\small
\begin{algorithmic}[1]
\Require Parallel data $\mathcal{P}$, language pool $\mathcal{L}_T$, inner steps $k$, temperature $\tau$
\State Initialize model parameters $\theta$, lang. weights $w \in \mathbb{R}^{|\mathcal{L}_T|}$
\For{each outer iteration $t$}
    \State $p \leftarrow \text{softmax}(w / \tau)$ \Comment{language sampling distribution}
    \State $L\in\mathbb{R}^{|\mathcal{L}_T|} \leftarrow \mathbf{0}$ \Comment{per-language loss accumulator}
    \For{$i = 1, \ldots, k$} \Comment{inner loop}
        \State Sample batch $\widetilde{\mathcal{P}}_i$ from $\mathcal{P}$ according to $p$
        \State $\theta \leftarrow \theta - \nabla_\theta L_R(\theta;\, \widetilde{\mathcal{P}}_i)$ \Comment{Eq.~\ref{eq:realignment}}
        \State Accumulate language contributions to $L_R$ into $L$
    \EndFor
    \State $w \leftarrow \text{updateWeights}(L, w)$ \Comment{variant-specific}
\EndFor
\end{algorithmic}
\end{algorithm}

We hypothesize that realignment requires more frequent exposure to ``difficult''
languages (linguistically distant or low-resource), and that these
requirements shift as training progresses. Thus, we introduce a bi-level optimization that adaptively samples languages depending on the training progression during realignment. Our framework is summarized in Algorithm~\ref{alg:bilevel}. We assign a scalar weight $w_\ell$
to each $\ell \in \mathcal{L}_T$, collected in $w \in
\mathbb{R}^{|\mathcal{L}_T|}$, inducing a sampling distribution $p =
\text{softmax}(w/\tau)$ with temperature $\tau$. An inner loop optimizes
$L_R$ with respect to $\theta$, sampling languages according to $p$; every $k$
inner steps, an outer loop updates $w$ from the accumulated per-language loss
signal so as to \emph{maximize} $L_R$, focusing sampling on languages that
remain hard to align.

\paragraph{UCB-based re-weighting.}
We frame language selection as a multi-armed bandit, each language an arm whose
reward is its per-language realignment loss~\citep{auer2002finite}. Let $n_\ell$
be the number of times $\ell$ has been sampled and $\bar{r}_\ell$ its mean
reward; at outer iteration $t$,
\begin{equation}
    \text{UCB}_\ell(t) = \bar{r}_\ell + c \sqrt{\frac{\ln t}{n_\ell + 1}},
\end{equation}
where $c > 0$ controls exploration. The first term exploits high-loss
(under-aligned) languages; the second explores infrequently sampled ones.
Unsampled languages receive a score above the current maximum to guarantee
initial coverage. We set $w_\ell \leftarrow \text{UCB}_\ell(t)$.

\paragraph{Gradient-based variant.}
As an alternative, $w$ can be updated by gradient ascent on the weighted
outer-loop loss $\sum_i w_i L_i$, trading UCB's explicit exploration term for
a smooth, differentiable update (details in Appendix~\ref{appendix:gradient}).

\felixremove{
\paragraph{Language Weighting.}

Given the language pool $\mathcal{L}_T$, we define a meta-weight vector $w \in \mathbb{R}^{|\mathcal{L}_T|}$ with one weight assigned to each language $\ell \in \mathcal{L}_T$. These weights induce a learnable probability distribution $p = \text{softmax}\left(\frac{w}{\tau}\right)$ over the target languages, with temperature $\tau$ controlling the "sharpness" of the probability distribution. For each target language $\ell$, its corresponding probability is denoted by $p(\ell)$, which determines the sampling of each language from the parallel data, while the weight will be optimized along with the realignment loss. We implement an outer-loop to optimize this loss after every $k$ iteration of realignment batches.

\paragraph{Outer-loop Loss. } For each batch, the outer-loop loss is the weighted average of \felix{the terms of the negative of the realignment loss $\mathcal{L}_R$ in Equation \ref{eq:realignment} where each term is scaled with a weight attributed to each language}.  \dav{Can you define what is $\mathcal{P}$? Also quite unclear why do you need to do summation w.r.t. $\ell_s$ and $\ell$, I thought these are languages? What does this represent, string or number?} \felixcomment{I agree that the notation is confusing, $\mathcal{P}$ was introduced in 2.1 and $\ell_S$ is the loss for language $\ell_S$, we're using the same letter with different writing... I've tried to clarify that below, but I still have an issue with $t$ being either a target sample ($(s,t)$ in $\mathcal{P}$) or the index of a target language (e.g. in $l_t$)}

\begin{equation}
    L_{\text{outer\_batch}} = - \frac{\sum_{(x,y, \ell) \in \widetilde{\mathcal{P}}} \left[ L(x,y) + L(y,x)\right] p(\ell)}{2 \sum_{(x,y, \ell) \in \widetilde{\mathcal{P}}} p(\ell)}
\end{equation}

\felix{Where $\mathcal{L}(x,y)$ and $\mathcal{L}(y,x)$ are the two loss terms already defined in Equation \ref{eq:realignment} for each sample $(x,y,\ell)$ in the realignment batch $\widetilde{\mathcal{P}}$ augmented with the target language  $\ell$ of sample $y$.} In total, the outer-loop loss for all languages will be calculated after $k$ realignment batches will be:

\begin{equation}
    L_{\text{outer\_total}} = \frac{1}{k} \sum_{i=1}^{k} L_{\text{outer\_batch}, i}
\end{equation}

\felix{$w$ is trained to minimize this outer-loop loss and thus to maximize the contribution of languages with a higher realignment loss. In other words, the outer-loop training tends to oversample languages that have a higher realignment loss.} \dav{It's not really clear how these relate with equation 1 or 2? Why would this work, what's the reasoning behind it? What are you minimizing/optimizing?}

The objective of this loss is to adaptively adjust the sampling distribution over languages by prioritizing hard-to-align languages - those that contribute most to the alignment loss - while down-weighting well-aligned languages, which are typically typologically closer to the pivot language. This sampling distribution is allowed to evolve throughout training. By focusing more on challenging language pairs, we hypothesize that the model can achieve improved cross-lingual alignment, thereby enhancing the transfer capability of multilingual language models.
}

\section{Experimental Setup}

\paragraph{Models.} Our main object of study is \textbf{XLM-R Large}
\citep{conneau-etal-2020-unsupervised}, a 560M-parameter encoder-only model
pretrained on 100 languages. To test generalization to decoder-only
models, we additionally include \textbf{Gemma~2~9B}
\citep{gemmateam2024gemma2improvingopen}, trained with LoRA adapters
\citep{hu2022lora} due to resource constraints, following
\citet{liu-niehues-2025-middle}. We discuss the rationale for focusing on
encoder-only models and classification tasks in Appendix~\ref{appendix:scope} and justification for using LoRA for generative models in Appendix~\ref{sec:lora_usage}.

We also perform a post-hoc analysis on models that weren't involved in the other experiments, to demonstrate the off-the-shelf effectiveness of our UCB-based approach (Section~\ref{sec:rebuttal}). We include three additional encoder-only models and one decoder-only model: \textbf{mBERT} \citep{bert}, \textbf{mDeBERTa v3} \citep{he2021debertav3}, \textbf{mmBERT} \citep{marone2026mmbert} and \textbf{Llama 3.1 8B} \citep{grattafiori2024llama3herd}. These four models constitute the complete set considered for this analysis. For the encoder-only models, we replicate the XLM-R Large training configuration. For Llama 3.1 8B, we mirror the Gemma 2 9B settings by employing LoRA adapters. For the gradient-based experiment, we used the outer-loop learning rate of $10^{-3}$ for all models.

\paragraph{Data and tasks.} We follow \citet{nguyen-etal-2025-rethinking}, evaluating on NER, POS, and NLI across 65 languages (29 low-resource), with parallel realignment data from OPUS-100 and NLLB. Additional dataset details and random seeds are provided in Appendix~\ref{appendix:data}.

\begin{table*}
\centering

\small
\adjustbox{max width=0.8\linewidth}{
\begin{tabularx}{\textwidth}{lXXXX}
\toprule
EXP & NER (F1) & NLI (Acc) & POS (Acc) & AVG \\
\midrule
\multicolumn{5}{l}{\textbf{XLM-R-Large}} \\
Fine-tuning only & 57.83 \textsubscript{$\pm$0.90} & 65.54 \textsubscript{$\pm$0.39} & 68.90 \textsubscript{$\pm$0.61} & 64.09 \textsubscript{$\pm$0.38} \\
Uniform realignment & 61.97 \textsubscript{$\pm$0.57} (+4.14) & 67.70 \textsubscript{$\pm$0.34} (+2.16) & 71.56 \textsubscript{$\pm$0.40} (+2.67) & 67.08 \textsubscript{$\pm$0.13} (+2.99) \\
Most-URIEL & 62.00 \textsubscript{$\pm$0.56} (+4.17) & 67.98 \textsubscript{$\pm$0.32} (+2.44) & 70.93 \textsubscript{$\pm$0.35} (+2.03) & 66.97 \textsubscript{$\pm$0.37} (+2.88) \\
Gradient-based & \textbf{62.25 \textsubscript{$\pm$1.10} (+4.42)} & 68.65 \textsubscript{$\pm$0.29} (+3.11) & \textbf{72.33 \textsubscript{$\pm$0.34} (+3.44)} & \textbf{67.75 \textsubscript{$\pm$0.31} (+3.66)} \\
UCB-based & 61.93 \textsubscript{$\pm$0.79} (+4.09) & \textbf{68.94 \textsubscript{$\pm$0.02} (+3.40)} & 72.33 \textsubscript{$\pm$0.42} (+3.44) & 67.73 \textsubscript{$\pm$0.13} (+3.64) \\
\midrule
\multicolumn{5}{l}{\textbf{Gemma-2-9b}} \\
Fine-tuning only & 36.09 \textsubscript{$\pm$1.18} & 63.30 \textsubscript{$\pm$3.67} & 49.09 \textsubscript{$\pm$0.94} & 49.49 \textsubscript{$\pm$1.42} \\
Uniform realignment & 36.51 \textsubscript{$\pm$1.35} (+0.42) & 69.97 \textsubscript{$\pm$1.08} (+6.67) & 48.22 \textsubscript{$\pm$4.70} (-0.87) & 51.57 \textsubscript{$\pm$2.24} (+2.07) \\
Most-URIEL & \textbf{37.60 \textsubscript{$\pm$1.17} (+1.51)} & \textbf{70.11 \textsubscript{$\pm$1.01}} (+6.81) & \textbf{51.83 \textsubscript{$\pm$0.85} (+2.74)} & \textbf{53.18 \textsubscript{$\pm$0.28} (+3.69)} \\
Gradient-based & 34.54 \textsubscript{$\pm$3.94} (-1.55) & 69.31 \textsubscript{$\pm$0.81} (+6.01) & 50.15 \textsubscript{$\pm$1.41} (+1.06) & 51.33 \textsubscript{$\pm$1.45} (+1.84) \\
UCB-based & 37.06 \textsubscript{$\pm$0.30} (+0.97) & 69.36 \textsubscript{$\pm$0.36} (+6.07) & 50.06 \textsubscript{$\pm$5.97} (+0.97) & 52.16 \textsubscript{$\pm$1.87} (+2.67) \\
\bottomrule
\end{tabularx}
}
\caption{Average performance across three random seeds for each task and across all three tasks, comparing three baselines with two dynamic sampling strategies. All realignment experiments use 16{,}000 realignment steps.}
\label{tab:res_overview}
\end{table*}

\paragraph{Baselines.} We compare against \textbf{Fine-tuning only} (no
realignment), \textbf{Uniform realignment}, and \textbf{Most-URIEL}
\citep{littell-etal-2017-uriel}, the 40 most typologically diverse languages
from \citet{nguyen-etal-2025-rethinking}. Our methods are \textbf{UCB-based}
and \textbf{Gradient-based}, all over the same 64 languages. Hyper-parameters
were fixed via a small grid search on XLM-R base
(Appendix~\ref{appendix:hyperparams}).

\section{Results}\label{results}

\subsection{Results Overview}

Table \ref{tab:res_overview} reports the cross-lingual performance gains of four realignment experiments compared to fine-tuning-only for 2 different model types. All realignment methods improve on the fine-tuning-only baseline for both
models, confirming the general effectiveness of realignment. Besides Uniform realignment for POS and gradient-based sampling for NER with Gemma, all realignment methods bring a performance gain ranging from $0.42$ to $6.81$ points, depending on the method and task.

On XLM-R, both dynamic methods (UCB and gradient-based) outperform the static baselines (Uniform, Most-URIEL). On average, the gradient-based approach achieves the highest gain, at the same time outperforming both uniform sampling and most-URIEL baseline, indicating that the dynamic sampling strategies are more effective than static subset selection or uniform sampling. These gains are consistent over tasks, with the largest improvements observed in NLI, with the UCB-based method achieving a gain of $+1.24$ points over uniform sampling.

On Gemma, the gradient-based method largely fails to generalize, whereas UCB still achieves positive gains across all three downstream tasks, suggesting that the gradient-based approach is more brittle across architectures. Unlike XLM-R, the realignment gains on Gemma are concentrated mainly on NLI tasks, reaching up to $6.81$ points of improvement. In some cases, realignment even harms POS tagging and NER performance, although the decoder-only architecture of Gemma is not well-suited for these token-level tasks as it cannot infer a token class from future tokens while encoder-only models can \citep{dukic-snajder-2024-looking}.

All realignment methods achieve significant gains on XNLI for Gemma, at least $6$ points, while the gap between the different sampling approaches is smaller than their standard deviation. This suggests that for Gemma, realignment itself is highly effective. Furthermore, the UCB-based method demonstrates strong robustness across architectural changes, performing well not only in the transition from encoder-only to decoder-only models, but also under a different training regime that uses adapters instead of full-model fine-tuning.

\subsection{Resource-level Break Down}

\begin{table}
\adjustbox{max width=\linewidth}{
\begin{tabular}{lcccc}
\toprule
type & HRL & MRL & LRL$_\text{seen}$ & LRL$_\text{unseen}$ \\
\midrule
Fine-tuning only & \textbf{75.86}$_{\pm 0.25}$ & \textbf{77.47}$_{\pm 0.21}$ & 55.03$_{\pm 0.92}$ & 43.74$_{\pm 1.01}$ \\
Uniform realignment & 75.03$_{\pm 0.62}$ & 76.87$_{\pm 0.16}$ & 55.88$_{\pm 0.72}$ & 53.13$_{\pm 0.10}$ \\
Most-URIEL & 74.96$_{\pm 0.08}$ & 76.95$_{\pm 0.09}$ & \textbf{56.24}$_{\pm 0.50}$ & 52.58$_{\pm 0.89}$ \\
Gradient-based & 75.14$_{\pm 0.92}$ & 77.05$_{\pm 0.18}$ & 55.51$_{\pm 0.75}$ & \textbf{54.74}$_{\pm 0.27}$ \\
UCB-based & 75.02$_{\pm 0.31}$ & 77.23$_{\pm 0.05}$ & 55.65$_{\pm 0.21}$ & 54.54$_{\pm 0.19}$ \\
\bottomrule
\end{tabular}
}
\caption{Cross-lingual transfer performance of XLM-R Large in different resource-level language groups.}
\label{tab:resource-level}
\end{table}

Table \ref{tab:resource-level} further breaks down models performance across different resource-level groups, including high-resource languages (HRLs), middle-resource languages (MRLs), low-resource languages (LRLs) seen and unseen. 
The exact language composition of each group is provided in Appendix~\ref{appendix:data}. The distinction between seen and unseen LRLs is not made for Gemma, as it is not reported which languages were included in its pre-training data.

Compared to static or uniform sampling, dynamic sampling distributes gains more evenly across resource levels for XLM-R. Most importantly, dynamic sampling mitigates the performance drop on MRLs, where uniform sampling triggers a $0.6$ point drop over simple fine-tuning, while UCB-based reduces this drop to only
$0.3$. Furthermore, XLM-R results on unseen languages demonstrate stronger generalization ability compared to the two realignment baselines, suggesting that dynamic sampling leads to more robust and generalized cross-lingual transfer. Previous literature has shown that realignment can be harmful to some
languages \citep{wu-dredze-2020-explicit,gaschi-etal-2023-exploring}, our results suggest that this is especially the case for MRLs and HRLs, and that dynamic sampling can mitigate this issue.

\subsection{Win Rate vs.\ Average Gap}

\begin{table}[t]
\centering
\small
\begin{tabular}{lcc}
\toprule
 & XLM-R & Gemma \\
\midrule
Avg.\ gain UCB $-$ Uniform & $+0.67$ & $+0.60$ \\
Win rate UCB $>$ Uniform & $64.2\%$ & $51.0\%$ \\
\bottomrule
\end{tabular}
\caption{Average accuracy gap and language-level win rate of UCB-based over uniform realignment (N=441 task--language--seed triplets).}
\label{tab:winrate}
\end{table}

The average gains in Table~\ref{tab:res_overview} understate the impact of
dynamic sampling. Table~\ref{tab:winrate} contrasts the average gap with the
language-level win rate, computed across all (task, language, seed) triplets.
On XLM-R, UCB only improves the macro-average by $+0.65$ points, but it
outperforms uniform realignment on $64.2\%$ of individual triplets, indicating
that the gains are spread across the benchmark rather than concentrated on a
few languages. On Gemma the effect is smaller (win rate $51.0\%$). The full pairwise win rate
matrix across all methods and both models, together with a detailed analysis,
is provided in Appendix~\ref{appendix:winloss}.

\subsection{Decoupling Language Weights from Dynamic Adaptation}

\begin{figure}
    \centering
    \includegraphics[width=\linewidth]{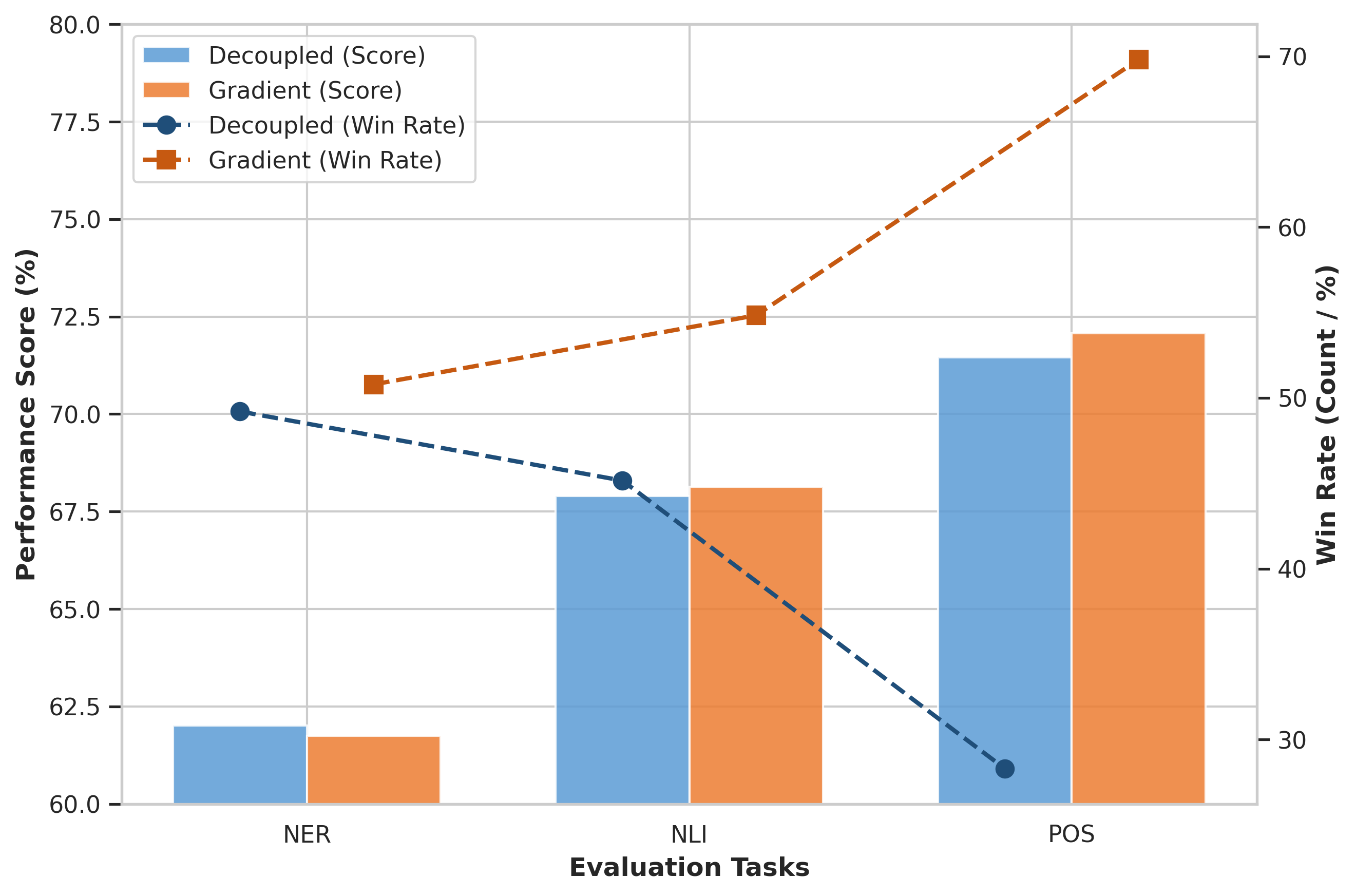}
    \caption{Performance and Win-rate comparison between the gradient-based method and the decoupled variant using seed 42 with 32{,}000 realignment steps.}
    \label{fig:decouple}
\end{figure}

\begin{table*}[t]
\centering
\small
\begin{tabular}{l c c c c c}
\hline
 & \textbf{Fine-tuning} & \textbf{Uniform} & \textbf{Most URIEL} & \textbf{UCB-based} & \textbf{Gradient-based}\\
 & \textbf{Only} & & & \textbf{(Ours)} & \textbf{(Ours)}\\
\hline
mBERT & 56.27 \textsubscript{$\pm$0.09} & \textbf{61.18} \textsubscript{$\pm$0.08} & 60.70 \textsubscript{$\pm$0.15} & \underline{61.17} \textsubscript{$\pm$0.25} & 61.05 \textsubscript{$\pm$0.25} \\
mDeBERTa v3 & 66.36 \textsubscript{$\pm$0.30} & 66.83 \textsubscript{$\pm$0.35} & 66.09 \textsubscript{$\pm$0.03} & \textbf{66.96} \textsubscript{$\pm$0.21} & \underline{66.85} \textsubscript{$\pm$0.25} \\
mmBERT & 54.72 \textsubscript{$\pm$1.30} & 61.21 \textsubscript{$\pm$0.78} & 61.01 \textsubscript{$\pm$0.46} & \underline{61.48} \textsubscript{$\pm$0.76} & \textbf{61.91} \textsubscript{$\pm$0.70} \\
Llama 3.1 8B & 42.30 \textsubscript{$\pm$0.33} & 45.10 \textsubscript{$\pm$2.92} & \underline{45.48} \textsubscript{$\pm$1.52} & 45.20 \textsubscript{$\pm$2.41} & \textbf{45.80} \textsubscript{$\pm$1.36} \\
\hdashline 
XLM-R Large & 64.09 \textsubscript{$\pm$0.38} & 67.08 \textsubscript{$\pm$0.13} & 66.97 \textsubscript{$\pm$0.37} & \underline{67.73} \textsubscript{$\pm$0.13} & \textbf{67.75} \textsubscript{$\pm$0.31} \\
Gemma-2-9b & 49.49 \textsubscript{$\pm$1.42} & 51.57 \textsubscript{$\pm$2.24} & \textbf{53.18} \textsubscript{$\pm$0.28} & 51.33 \textsubscript{$\pm$1.87} & \underline{52.16} \textsubscript{$\pm$1.45} \\
\hline
\end{tabular}
\caption{Macro-average performance of additional encoder-only and decoder-only models on the same three tasks with the same setting and three seeds as XLM-R and Gemma 2. Bold indicates best, and underline indicates second-best in a row. The bottom two rows, separated by the dashed line, are imported from Table~\ref{tab:res_overview} for direct comparison. Details per task can be found in Appendix Table~\ref{tab:res_all}.}
\label{tab:additional_models}
\end{table*}

A natural question is whether the gains stem from discovering a better language distribution or from the dynamic co-adaptation of sampling and model parameters. To disentangle these effects, we conduct a decoupling ablation on \texttt{XLM-R-Large}: we first learn language weights with the model frozen, then freeze those weights and run realignment using this fixed distribution, followed by standard fine-tuning.

Figure~\ref{fig:decouple} shows that although the downstream task performance differences are relatively small, the gradient-based method achieves better task-wise win--loss rates and more generalized cross-lingual transfer compared to the decoupled variant. This suggests that the benefit of dynamic sampling does not primarily come from converging to a single optimal distribution, but rather from the co-evolution of language weights and model representations throughout training. As realignment reshapes the embedding space, the relative alignment difficulty of each language also changes, requiring the sampling strategy to adapt dynamically. In particular, language weights that are effective during the early stages of training may become suboptimal later on. As a result, any fixed distribution, even one initially learned from the loss signal, may eventually become misaligned with the model’s evolving training needs.

\subsection{Robustness across different models}
\label{sec:rebuttal}

To demonstrate the robustness of our approach across different models, we extend our evaluation to include three additional encoder-only models (mBERT, mDeBERTa v3, and mmBERT) and one decoder-only model (Llama 3.1 8B).

Table~\ref{tab:additional_models} indicate that our two proposed approaches outperform both uniform sampling and standard fine-tuning across the majority of evaluated models. While the Most-URIEL baseline surpasses uniform sampling on decoder-only models, it consistently underperforms on encoder-only architectures. Furthermore, although Most-URIEL previously set a strong baseline on Gemma-2-9B, our Gradient-based strategy exceeds it, achieving the highest macro-average results on Llama 3.1 8B, yielding up to a 3.5\% point gain over fine-tuning alone. Overall, these results demonstrate that our proposed methods offer superior consistency across diverse model architectures and training recipes over uniform sampling or performing realignment on a fixed subset.

\section{Conclusion}

Our work introduces two dynamic sampling strategies for parallel data selection during realignment that yield more generalized improvements on \texttt{XLM-R-Large}. Furthermore, we extend and validate the benefits of realignment for decoder-only models in multilingual natural language understanding. Lastly, we find that average accuracy understates the advantage of dynamic sampling: gains are more evenly distributed across languages, improving or maintaining LRL results while mitigating the MRL accuracy drop that realignment can trigger. Among our approaches, UCB-based sampling shows greater robustness to training duration and hyperparameter choice, requiring no per-model tuning to remain competitive, whereas gradient-based strategies can achieve larger gains on some architectures (e.g. mmBERT, Llama 3.1 8B) but need model-wise configuration tuning and are more prone to degrading outside their tuned regime. We hope this work encourages further research on data utilization strategies for continual multilingual training and cross-lingual transfer.

\section*{Limitations}

The conclusions of this work are limited by the scope of our experiments, which
focus on encoder-only models and classification tasks, due to a limited
availability of resources, both computational and data-related. The compute we
had only allowed for realignment with adapters on small LLMs (9B model), and the
lack of large-scale multilingual generative benchmarks covering a wide range of
LRLs motivated us to focus on classification tasks. Although we include
results for a decoder-only model, a more comprehensive evaluation on generative
tasks would be a valuable complement to this work.

Due to a lack of available human speakers of the studied low-resource languages,
and the large scale of our experiments, we only evaluate on quantitative
metrics, which might not fully capture the quality of the realignment and its
downstream impact. Future work could focus on a smaller set of languages and
include a more qualitative evaluation, especially regarding generative tasks,
given the limitations of quantitative metrics for generative tasks.

Because we can only evaluate on the languages for which we have annotated data
in at least one downstream task, our evaluation is limited to 78 languages,
which is a small fraction of the world's languages. Although this is a sizable
coverage, it still limits the generalizability of our conclusions, and presents
the risk of overexposure to certain language families and typological features. 

\bibliography{colm2026_conference}

@inproceedings{gaschi-etal-2023-exploring,
    title = "Exploring the Relationship between Alignment and Cross-lingual Transfer in Multilingual Transformers",
    author = "Gaschi, Felix  and
      Cerda, Patricio  and
      Rastin, Parisa  and
      Toussaint, Yannick",
    editor = "Rogers, Anna  and
      Boyd-Graber, Jordan  and
      Okazaki, Naoaki",
    booktitle = "Findings of the Association for Computational Linguistics: ACL 2023",
    month = jul,
    year = "2023",
    address = "Toronto, Canada",
    publisher = "Association for Computational Linguistics",
    url = "https://aclanthology.org/2023.findings-acl.189/",
    doi = "10.18653/v1/2023.findings-acl.189",
    pages = "3020--3042"
}

@inproceedings{nguyen-etal-2025-rethinking,
    title = "Rethinking what matters: Effective and Robust Multilingual Realignment for Low-Resource Languages",
    author = "Nguyen, Quang Phuoc  and
      Anugraha, David  and
      Gaschi, F{\'e}lix  and
      Cheng, Jun Bin  and
      Lee, En-Shiun Annie",
    editor = "Inui, Kentaro  and
      Sakti, Sakriani  and
      Wang, Haofen  and
      Wong, Derek F.  and
      Bhattacharyya, Pushpak  and
      Banerjee, Biplab  and
      Ekbal, Asif  and
      Chakraborty, Tanmoy  and
      Singh, Dhirendra Pratap",
    booktitle = "Proceedings of the 14th International Joint Conference on Natural Language Processing and the 4th Conference of the Asia-Pacific Chapter of the Association for Computational Linguistics",
    month = dec,
    year = "2025",
    address = "Mumbai, India",
    publisher = "The Asian Federation of Natural Language Processing and The Association for Computational Linguistics",
    url = "https://aclanthology.org/2025.ijcnlp-long.102/",
    pages = "1877--1905",
    ISBN = "979-8-89176-298-5"
}

@inproceedings{wu-dredze-2020-explicit,
    title = "Do Explicit Alignments Robustly Improve Multilingual Encoders?",
    author = "Wu, Shijie  and
      Dredze, Mark",
    editor = "Webber, Bonnie  and
      Cohn, Trevor  and
      He, Yulan  and
      Liu, Yang",
    booktitle = "Proceedings of the 2020 Conference on Empirical Methods in Natural Language Processing (EMNLP)",
    month = nov,
    year = "2020",
    address = "Online",
    publisher = "Association for Computational Linguistics",
    url = "https://aclanthology.org/2020.emnlp-main.362/",
    doi = "10.18653/v1/2020.emnlp-main.362",
    pages = "4471--4482"
}

@inproceedings{kargaran-etal-2025-mexa,
    title = "{MEXA}: Multilingual Evaluation of {E}nglish-Centric {LLM}s via Cross-Lingual Alignment",
    author = "Kargaran, Amir Hossein  and
      Modarressi, Ali  and
      Nikeghbal, Nafiseh  and
      Diesner, Jana  and
      Yvon, Fran{\c{c}}ois  and
      Schuetze, Hinrich",
    editor = "Che, Wanxiang  and
      Nabende, Joyce  and
      Shutova, Ekaterina  and
      Pilehvar, Mohammad Taher",
    booktitle = "Findings of the Association for Computational Linguistics: ACL 2025",
    month = jul,
    year = "2025",
    address = "Vienna, Austria",
    publisher = "Association for Computational Linguistics",
    url = "https://aclanthology.org/2025.findings-acl.1385/",
    doi = "10.18653/v1/2025.findings-acl.1385",
    pages = "27001--27023",
    ISBN = "979-8-89176-256-5"
}

@inproceedings{liu-niehues-2025-middle,
    title = "Middle-Layer Representation Alignment for Cross-Lingual Transfer in Fine-Tuned {LLM}s",
    author = "Liu, Danni  and
      Niehues, Jan",
    editor = "Che, Wanxiang  and
      Nabende, Joyce  and
      Shutova, Ekaterina  and
      Pilehvar, Mohammad Taher",
    booktitle = "Proceedings of the 63rd Annual Meeting of the Association for Computational Linguistics (Volume 1: Long Papers)",
    month = jul,
    year = "2025",
    address = "Vienna, Austria",
    publisher = "Association for Computational Linguistics",
    url = "https://aclanthology.org/2025.acl-long.778/",
    doi = "10.18653/v1/2025.acl-long.778",
    pages = "15979--15996",
    ISBN = "979-8-89176-251-0"
}

@inproceedings{bakos-etal-2025-alignfreeze,
    title = "{A}lign{F}reeze: Navigating the Impact of Realignment on the Layers of Multilingual Models Across Diverse Languages",
    author = "Bakos, Steve  and
      Guzm{\'a}n, David  and
      More, Riddhi  and
      Li, Kelly Chutong  and
      Gaschi, F{\'e}lix  and
      Lee, En-Shiun Annie",
    editor = "Chiruzzo, Luis  and
      Ritter, Alan  and
      Wang, Lu",
    booktitle = "Proceedings of the 2025 Conference of the Nations of the Americas Chapter of the Association for Computational Linguistics: Human Language Technologies (Volume 2: Short Papers)",
    month = apr,
    year = "2025",
    address = "Albuquerque, New Mexico",
    publisher = "Association for Computational Linguistics",
    url = "https://aclanthology.org/2025.naacl-short.48/",
    doi = "10.18653/v1/2025.naacl-short.48",
    pages = "562--586",
    ISBN = "979-8-89176-190-2"
}

@inproceedings{chen-etal-2025-scale,
    title = "Scale Down to Speed Up: Dynamic Data Selection for Reinforcement Learning",
    author = "Chen, Zhuoyue  and
      Zhang, Jihai  and
      Liu, Ben  and
      Lin, Fangquan  and
      Yin, Wotao",
    editor = "Christodoulopoulos, Christos  and
      Chakraborty, Tanmoy  and
      Rose, Carolyn  and
      Peng, Violet",
    booktitle = "Findings of the Association for Computational Linguistics: EMNLP 2025",
    month = nov,
    year = "2025",
    address = "Suzhou, China",
    publisher = "Association for Computational Linguistics",
    url = "https://aclanthology.org/2025.findings-emnlp.412/",
    doi = "10.18653/v1/2025.findings-emnlp.412",
    pages = "7806--7817",
    ISBN = "979-8-89176-335-7"
}

@misc{xie2023doremioptimizingdatamixtures,
      title={DoReMi: Optimizing Data Mixtures Speeds Up Language Model Pretraining}, 
      author={Sang Michael Xie and Hieu Pham and Xuanyi Dong and Nan Du and Hanxiao Liu and Yifeng Lu and Percy Liang and Quoc V. Le and Tengyu Ma and Adams Wei Yu},
      year={2023},
      eprint={2305.10429},
      archivePrefix={arXiv},
      primaryClass={cs.CL},
      url={https://arxiv.org/abs/2305.10429}, 
}

@misc{albalak2023efficientonlinedatamixing,
      title={Efficient Online Data Mixing For Language Model Pre-Training}, 
      author={Alon Albalak and Liangming Pan and Colin Raffel and William Yang Wang},
      year={2023},
      eprint={2312.02406},
      archivePrefix={arXiv},
      primaryClass={cs.CL},
      url={https://arxiv.org/abs/2312.02406}, 
}

@InProceedings{pmlr-v70-finn17a,
  title = 	 {Model-Agnostic Meta-Learning for Fast Adaptation of Deep Networks},
  author =       {Chelsea Finn and Pieter Abbeel and Sergey Levine},
  booktitle = 	 {Proceedings of the 34th International Conference on Machine Learning},
  pages = 	 {1126--1135},
  year = 	 {2017},
  editor = 	 {Precup, Doina and Teh, Yee Whye},
  volume = 	 {70},
  series = 	 {Proceedings of Machine Learning Research},
  month = 	 {06--11 Aug},
  publisher =    {PMLR},
  url = 	 {https://proceedings.mlr.press/v70/finn17a.html}
}

@misc{nichol2018firstordermetalearningalgorithms,
      title={On First-Order Meta-Learning Algorithms}, 
      author={Alex Nichol and Joshua Achiam and John Schulman},
      year={2018},
      eprint={1803.02999},
      archivePrefix={arXiv},
      primaryClass={cs.LG},
      url={https://arxiv.org/abs/1803.02999}, 
}

@misc{ren2019learningreweightexamplesrobust,
      title={Learning to Reweight Examples for Robust Deep Learning}, 
      author={Mengye Ren and Wenyuan Zeng and Bin Yang and Raquel Urtasun},
      year={2019},
      eprint={1803.09050},
      archivePrefix={arXiv},
      primaryClass={cs.LG},
      url={https://arxiv.org/abs/1803.09050}, 
}

@inproceedings{joshi-etal-2020-state,
    title = "The State and Fate of Linguistic Diversity and Inclusion in the {NLP} World",
    author = "Joshi, Pratik  and
      Santy, Sebastin  and
      Budhiraja, Amar  and
      Bali, Kalika  and
      Choudhury, Monojit",
    editor = "Jurafsky, Dan  and
      Chai, Joyce  and
      Schluter, Natalie  and
      Tetreault, Joel",
    booktitle = "Proceedings of the 58th Annual Meeting of the Association for Computational Linguistics",
    month = jul,
    year = "2020",
    address = "Online",
    publisher = "Association for Computational Linguistics",
    url = "https://aclanthology.org/2020.acl-main.560/",
    doi = "10.18653/v1/2020.acl-main.560",
    pages = "6282--6293",
}

@inproceedings{pires-etal-2019-multilingual,
    title = "How Multilingual is Multilingual {BERT}?",
    author = "Pires, Telmo  and
      Schlinger, Eva  and
      Garrette, Dan",
    editor = "Korhonen, Anna  and
      Traum, David  and
      M{\`a}rquez, Llu{\'i}s",
    booktitle = "Proceedings of the 57th Annual Meeting of the Association for Computational Linguistics",
    month = jul,
    year = "2019",
    address = "Florence, Italy",
    publisher = "Association for Computational Linguistics",
    url = "https://aclanthology.org/P19-1493/",
    doi = "10.18653/v1/P19-1493",
    pages = "4996--5001",
}

@misc{cao2020multilingualalignmentcontextualword,
      title={Multilingual Alignment of Contextual Word Representations}, 
      author={Steven Cao and Nikita Kitaev and Dan Klein},
      year={2020},
      eprint={2002.03518},
      archivePrefix={arXiv},
      primaryClass={cs.CL},
      url={https://arxiv.org/abs/2002.03518}, 
}

@article{auer2002finite,
  author  = {Auer, Peter and Cesa-Bianchi, Nicolò and Fischer, Paul},
  title   = {Finite-time Analysis of the Multiarmed Bandit Problem},
  journal = {Machine Learning},
  year    = {2002},
  volume  = {47},
  pages   = {235--256},
  doi     = {10.1023/A:1013689704352}
}

@inproceedings{dukic-snajder-2024-looking,
    title = "Looking Right is Sometimes Right: Investigating the Capabilities of Decoder-only {LLM}s for Sequence Labeling",
    author = "Duki{\'c}, David  and
      {\v{S}}najder, Jan",
    editor = "Ku, Lun-Wei  and
      Martins, Andre  and
      Srikumar, Vivek",
    booktitle = "Findings of the Association for Computational Linguistics: ACL 2024",
    month = aug,
    year = "2024",
    address = "Bangkok, Thailand",
    publisher = "Association for Computational Linguistics",
    url = "https://aclanthology.org/2024.findings-acl.843/",
    doi = "10.18653/v1/2024.findings-acl.843",
    pages = "14168--14181",
}

@inproceedings{elshabrawy-etal-2025-statement,
    title = "Statement-Tuning Enables Efficient Cross-lingual Generalization in Encoder-only Models",
    author = "Elshabrawy, Ahmed  and
      Nguyen, Thanh-Nhi  and
      Kang, Yeeun  and
      Feng, Lihan  and
      Jain, Annant  and
      Shaikh, Faadil Abdullah  and
      Mansurov, Jonibek  and
      Imam, Mohamed Fazli Mohamed  and
      Ortiz-Barajas, Jesus-German  and
      Chevi, Rendi  and
      Aji, Alham Fikri",
    editor = "Che, Wanxiang  and
      Nabende, Joyce  and
      Shutova, Ekaterina  and
      Pilehvar, Mohammad Taher",
    booktitle = "Findings of the Association for Computational Linguistics: ACL 2025",
    month = jul,
    year = "2025",
    address = "Vienna, Austria",
    publisher = "Association for Computational Linguistics",
    url = "https://aclanthology.org/2025.findings-acl.835/",
    doi = "10.18653/v1/2025.findings-acl.835",
    pages = "16226--16248",
    ISBN = "979-8-89176-256-5",
}

@inproceedings{shcharbakova-etal-2025-scale,
    title = "When Scale Meets Diversity: Evaluating Language Models on Fine-Grained Multilingual Claim Verification",
    author = "Shcharbakova, Hanna  and
      Anikina, Tatiana  and
      Skachkova, Natalia  and
      Genabith, Josef Van",
    editor = "Akhtar, Mubashara  and
      Aly, Rami  and
      Christodoulopoulos, Christos  and
      Cocarascu, Oana  and
      Guo, Zhijiang  and
      Mittal, Arpit  and
      Schlichtkrull, Michael  and
      Thorne, James  and
      Vlachos, Andreas",
    booktitle = "Proceedings of the Eighth Fact Extraction and VERification Workshop (FEVER)",
    month = jul,
    year = "2025",
    address = "Vienna, Austria",
    publisher = "Association for Computational Linguistics",
    url = "https://aclanthology.org/2025.fever-1.5/",
    doi = "10.18653/v1/2025.fever-1.5",
    pages = "69--84",
    ISBN = "978-1-959429-53-1",
}

@inproceedings{boizard2025eurobert,
title={Euro{BERT}: Scaling Multilingual Encoders for European Languages},
author={Nicolas Boizard and Hippolyte Gisserot-Boukhlef and Duarte Miguel Alves and Andre Martins and Ayoub Hammal and Caio Corro and CELINE HUDELOT and Emmanuel Malherbe and Etienne Malaboeuf and Fanny Jourdan and Gabriel Hautreux and Jo{\~a}o Alves and Kevin El Haddad and Manuel Faysse and Maxime Peyrard and Nuno M Guerreiro and Patrick Fernandes and Ricardo Rei and Pierre Colombo},
booktitle={Second Conference on Language Modeling},
year={2025},
url={https://openreview.net/forum?id=jdOC24msVq}
}

@inproceedings{clark-etal-2020-tydi,
    title = "{T}y{D}i {QA}: A Benchmark for Information-Seeking Question Answering in Typologically Diverse Languages",
    author = "Clark, Jonathan H. and Choi, Eunsol and Collins, Michael and Garrette, Dan and Kwiatkowski, Tom and Nikolaev, Vitaly and Palomaki, Jennimaria",
    booktitle = "Transactions of the Association for Computational Linguistics",
    volume = "8",
    year = "2020",
    pages = "454--470",
    url = "https://aclanthology.org/2020.tacl-1.30",
}

@inproceedings{lewis-etal-2020-mlqa,
    title = "{MLQA}: Evaluating Cross-lingual Extractive Question Answering",
    author = "Lewis, Patrick and O{\v{g}}uz, Barı{\c{s}} and Rinott, Ruty and Riedel, Sebastian and Schwenk, Holger",
    booktitle = "Proceedings of the 58th Annual Meeting of the Association for Computational Linguistics",
    year = "2020",
    publisher = "Association for Computational Linguistics",
    url = "https://aclanthology.org/2020.acl-main.653",
    pages = "7315--7330",
}

@inproceedings{longpre-etal-2021-mkqa,
    title = "{MKQA}: A Linguistically Diverse Benchmark for Multilingual Open Domain Question Answering",
    author = "Longpre, Shayne and Lu, Yi and Daiber, Joachim",
    booktitle = "Transactions of the Association for Computational Linguistics",
    volume = "9",
    year = "2021",
    pages = "1389--1406",
    url = "https://aclanthology.org/2021.tacl-1.82",
}

@inproceedings{conneau-etal-2020-unsupervised,
    title = "Unsupervised Cross-lingual Representation Learning at Scale",
    author = "Conneau, Alexis  and
      Khandelwal, Kartikay  and
      Goyal, Naman  and
      Chaudhary, Vishrav  and
      Wenzek, Guillaume  and
      Guzm{\'a}n, Francisco  and
      Grave, Edouard  and
      Ott, Myle  and
      Zettlemoyer, Luke  and
      Stoyanov, Veselin",
    editor = "Jurafsky, Dan  and
      Chai, Joyce  and
      Schluter, Natalie  and
      Tetreault, Joel",
    booktitle = "Proceedings of the 58th Annual Meeting of the Association for Computational Linguistics",
    month = jul,
    year = "2020",
    address = "Online",
    publisher = "Association for Computational Linguistics",
    url = "https://aclanthology.org/2020.acl-main.747/",
    doi = "10.18653/v1/2020.acl-main.747",
    pages = "8440--8451",
}

@inproceedings{hu-etal-2020-xtreme,
    title = "{XTREME}: A Massively Multilingual Multi-task Benchmark for Evaluating Cross-lingual Generalisation",
    author = "Hu, Junjie and Ruder, Sebastian and Siddhant, Aditya and Neubig, Graham and Firat, Orhan and Johnson, Melvin",
    booktitle = "Proceedings of the 37th International Conference on Machine Learning",
    year = "2020",
    url = "https://arxiv.org/abs/2003.11080",
}

@misc{gemmateam2024gemma2improvingopen,
      title={Gemma 2: Improving Open Language Models at a Practical Size}, 
      author={Gemma Team and Morgane Riviere and Shreya Pathak and Pier Giuseppe Sessa and Cassidy Hardin and Surya Bhupatiraju and Léonard Hussenot and Thomas Mesnard and Bobak Shahriari and Alexandre Ramé and Johan Ferret and Peter Liu and Pouya Tafti and Abe Friesen and Michelle Casbon and Sabela Ramos and Ravin Kumar and Charline Le Lan and Sammy Jerome and Anton Tsitsulin and Nino Vieillard and Piotr Stanczyk and Sertan Girgin and Nikola Momchev and Matt Hoffman and Shantanu Thakoor and Jean-Bastien Grill and Behnam Neyshabur and Olivier Bachem and Alanna Walton and Aliaksei Severyn and Alicia Parrish and Aliya Ahmad and Allen Hutchison and Alvin Abdagic and Amanda Carl and Amy Shen and Andy Brock and Andy Coenen and Anthony Laforge and Antonia Paterson and Ben Bastian and Bilal Piot and Bo Wu and Brandon Royal and Charlie Chen and Chintu Kumar and Chris Perry and Chris Welty and Christopher A. Choquette-Choo and Danila Sinopalnikov and David Weinberger and Dimple Vijaykumar and Dominika Rogozińska and Dustin Herbison and Elisa Bandy and Emma Wang and Eric Noland and Erica Moreira and Evan Senter and Evgenii Eltyshev and Francesco Visin and Gabriel Rasskin and Gary Wei and Glenn Cameron and Gus Martins and Hadi Hashemi and Hanna Klimczak-Plucińska and Harleen Batra and Harsh Dhand and Ivan Nardini and Jacinda Mein and Jack Zhou and James Svensson and Jeff Stanway and Jetha Chan and Jin Peng Zhou and Joana Carrasqueira and Joana Iljazi and Jocelyn Becker and Joe Fernandez and Joost van Amersfoort and Josh Gordon and Josh Lipschultz and Josh Newlan and Ju-yeong Ji and Kareem Mohamed and Kartikeya Badola and Kat Black and Katie Millican and Keelin McDonell and Kelvin Nguyen and Kiranbir Sodhia and Kish Greene and Lars Lowe Sjoesund and Lauren Usui and Laurent Sifre and Lena Heuermann and Leticia Lago and Lilly McNealus and Livio Baldini Soares and Logan Kilpatrick and Lucas Dixon and Luciano Martins and Machel Reid and Manvinder Singh and Mark Iverson and Martin Görner and Mat Velloso and Mateo Wirth and Matt Davidow and Matt Miller and Matthew Rahtz and Matthew Watson and Meg Risdal and Mehran Kazemi and Michael Moynihan and Ming Zhang and Minsuk Kahng and Minwoo Park and Mofi Rahman and Mohit Khatwani and Natalie Dao and Nenshad Bardoliwalla and Nesh Devanathan and Neta Dumai and Nilay Chauhan and Oscar Wahltinez and Pankil Botarda and Parker Barnes and Paul Barham and Paul Michel and Pengchong Jin and Petko Georgiev and Phil Culliton and Pradeep Kuppala and Ramona Comanescu and Ramona Merhej and Reena Jana and Reza Ardeshir Rokni and Rishabh Agarwal and Ryan Mullins and Samaneh Saadat and Sara Mc Carthy and Sarah Cogan and Sarah Perrin and Sébastien M. R. Arnold and Sebastian Krause and Shengyang Dai and Shruti Garg and Shruti Sheth and Sue Ronstrom and Susan Chan and Timothy Jordan and Ting Yu and Tom Eccles and Tom Hennigan and Tomas Kocisky and Tulsee Doshi and Vihan Jain and Vikas Yadav and Vilobh Meshram and Vishal Dharmadhikari and Warren Barkley and Wei Wei and Wenming Ye and Woohyun Han and Woosuk Kwon and Xiang Xu and Zhe Shen and Zhitao Gong and Zichuan Wei and Victor Cotruta and Phoebe Kirk and Anand Rao and Minh Giang and Ludovic Peran and Tris Warkentin and Eli Collins and Joelle Barral and Zoubin Ghahramani and Raia Hadsell and D. Sculley and Jeanine Banks and Anca Dragan and Slav Petrov and Oriol Vinyals and Jeff Dean and Demis Hassabis and Koray Kavukcuoglu and Clement Farabet and Elena Buchatskaya and Sebastian Borgeaud and Noah Fiedel and Armand Joulin and Kathleen Kenealy and Robert Dadashi and Alek Andreev},
      year={2024},
      eprint={2408.00118},
      archivePrefix={arXiv},
      primaryClass={cs.CL},
      url={https://arxiv.org/abs/2408.00118}, 
}

@inproceedings{hu2022lora,
title={Lo{RA}: Low-Rank Adaptation of Large Language Models},
author={Edward J Hu and yelong shen and Phillip Wallis and Zeyuan Allen-Zhu and Yuanzhi Li and Shean Wang and Lu Wang and Weizhu Chen},
booktitle={International Conference on Learning Representations},
year={2022},
url={https://openreview.net/forum?id=nZeVKeeFYf9}
}

@inproceedings{ebrahimi-etal-2022-americasnli,
    title = "{A}mericas{NLI}: Evaluating Zero-shot Natural Language Understanding of Pretrained Multilingual Models in Truly Low-resource Languages",
    author = "Ebrahimi, Abteen and Mager, Manuel and Oncevay, Arturo and Chaudhary, Vishrav and Chiruzzo, Luis and Fan, Angela and Ortega, John and Rios, Annette and Giménez-Lugo, Gustavo A. and Meza Ruiz, Ivan Vladimir and Neubig, Graham and Palmer, Alexis and Coto-Solano, Rolando and Vu, Ngoc Thang and Kann, Katharina",
    booktitle = "Proceedings of the 60th Annual Meeting of the Association for Computational Linguistics (Volume 1: Long Papers)",
    year = "2022",
    publisher = "Association for Computational Linguistics",
    url = "https://aclanthology.org/2022.acl-long.442",
    pages = "6351--6377",
}

@inproceedings{littell-etal-2017-uriel,
    title = "{URIEL} and lang2vec: Representing languages as typological, geographical, and phylogenetic vectors",
    author = "Littell, Patrick and Mortensen, David R. and Lin, Ke and Kairis, Katherine and Turner, Carlisle and Levin, Lori",
    booktitle = "Proceedings of the 15th Conference of the {E}uropean Chapter of the Association for Computational Linguistics: Volume 2, Short Papers",
    month = apr,
    year = "2017",
    address = "Valencia, Spain",
    publisher = "Association for Computational Linguistics",
    url = "https://aclanthology.org/E17-2002/",
    pages = "8--14",
}

@inproceedings{adelani-etal-2022-masakhaner2,
    title = "{M}asakha{NER} 2.0: {A}frica-centric Transfer Learning for Named Entity Recognition",
    author = "Adelani, David Ifeoluwa and Neubig, Graham and Ruder, Sebastian and Rijhwani, Shruti and Beukman, Michael and Palen-Michel, Chester and Lignos, Constantine and Alabi, Jesujoba O. and Muhammad, Shamsuddeen H. and Nabende, Peter and Dione, Cheikh M. Bamba and Bukula, Andiswa and Mabuya, Rooweither and Dossou, Bonaventure F. P. and Sibanda, Blessing and Buzaaba, Happy and Mukiibi, Jonathan and Kalipe, Godson and Mbaye, Derguene and Taylor, Amelia and Kabore, Fatoumata and Emezue, Chris Chinenye and Aremu, Anuoluwapo and Ogayo, Perez and Gitau, Catherine and Munkoh-Buabeng, Edwin and Memdjokam Koagne, Victoire and Tapo, Allahsera Auguste and Macucwa, Tebogo and Marivate, Vukosi and Mboning, Elvis and Gwadabe, Tajuddeen and Adewumi, Tosin and Ahia, Orevaoghene and Nakatumba-Nabende, Joyce and Mokono, Neo L. and Ezeani, Ignatius and Chukwuneke, Chiamaka and Adeyemi, Mofetoluwa and Hacheme, Gilles Q. and Abdulmumin, Idris and Ogundepo, Odunayo and Yousuf, Oreen and Moteu Ngoli, Tatiana and Klakow, Dietrich",
    booktitle = "Proceedings of the 2022 Conference on Empirical Methods in Natural Language Processing",
    month = dec,
    year = "2022",
    address = "Abu Dhabi, United Arab Emirates",
    publisher = "Association for Computational Linguistics",
    url = "https://aclanthology.org/2022.emnlp-main.298/",
    doi = "10.18653/v1/2022.emnlp-main.298",
    pages = "4488--4508",
}

@inproceedings{dione-etal-2023-masakhapos,
    title = "{M}asakha{POS}: Part-of-Speech Tagging for Typologically Diverse {A}frican languages",
    author = "Dione, Cheikh M. Bamba and Adelani, David Ifeoluwa and Nabende, Peter and Alabi, Jesujoba and Sindane, Thapelo and Buzaaba, Happy and Muhammad, Shamsuddeen Hassan and Emezue, Chris Chinenye and Ogayo, Perez and Aremu, Anuoluwapo and Gitau, Catherine and Mbaye, Derguene and Mukiibi, Jonathan and Sibanda, Blessing and Dossou, Bonaventure F. P. and Bukula, Andiswa and Mabuya, Rooweither and Tapo, Allahsera Auguste and Munkoh-Buabeng, Edwin and Memdjokam Koagne, Victoire and Ouoba Kabore, Fatoumata and Taylor, Amelia and Kalipe, Godson and Macucwa, Tebogo and Marivate, Vukosi and Gwadabe, Tajuddeen and Elvis, Mboning Tchiaze and Onyenwe, Ikechukwu and Atindogbe, Gratien and Adelani, Tolulope and Akinade, Idris and Samuel, Olanrewaju and Nahimana, Marien and Musabeyezu, Théogène and Niyomutabazi, Emile and Chimhenga, Ester and Gotosa, Kudzai and Mizha, Patrick and Agbolo, Apelete and Traore, Seydou and Uchechukwu, Chinedu and Yusuf, Aliyu and Abdullahi, Muhammad and Klakow, Dietrich",
    booktitle = "Proceedings of the 61st Annual Meeting of the Association for Computational Linguistics (Volume 1: Long Papers)",
    month = jul,
    year = "2023",
    address = "Toronto, Canada",
    publisher = "Association for Computational Linguistics",
    url = "https://aclanthology.org/2023.acl-long.609/",
    doi = "10.18653/v1/2023.acl-long.609",
    pages = "10883--10900",
}

@inproceedings{adelani-etal-2024-irokobench,
    title = "{I}roko{B}ench: A New Benchmark for {A}frican Languages in the Age of Large Language Models",
    author = "Adelani, David Ifeoluwa and Ojo, Jessica and Azime, Israel Abebe and Zhuang, Jian Yun and Alabi, Jesujoba Oluwadara and He, Xuanli and Ochieng, Millicent and Hooker, Sara and Bukula, Andiswa and Lee, En-Shiun Annie and Chukwuneke, Chiamaka Ijeoma and Buzaaba, Happy and Sibanda, Blessing Kudzaishe and Kalipe, Godson Koffi and Mukiibi, Jonathan and Kabongo Kabenamualu, Salomon and Yuehgoh, Foutse and Setaka, Mmasibidi and Ndolela, Lolwethu and Odu, Nkiruka and Mabuya, Rooweither and Osei, Salomey and Muhammad, Shamsuddeen Hassan and Samb, Sokhar and Guge, Tadesse Kebede and Sherman, Tombekai Vangoni and Stenetorp, Pontus",
    booktitle = "Proceedings of the 2025 Conference of the Nations of the Americas Chapter of the Association for Computational Linguistics: Human Language Technologies (Volume 1: Long Papers)",
    month = apr,
    year = "2025",
    address = "Albuquerque, New Mexico",
    publisher = "Association for Computational Linguistics",
    url = "https://aclanthology.org/2025.naacl-long.139/",
    doi = "10.18653/v1/2025.naacl-long.139",
    pages = "2732--2757",
}

@article{conneau2018xnli,
  title={XNLI: Evaluating cross-lingual sentence representations},
  author={Conneau, Alexis and Lample, Guillaume and Rinott, Ruty and Williams, Adina and Bowman, Samuel R and Schwenk, Holger and Stoyanov, Veselin},
  journal={arXiv preprint arXiv:1809.05053},
  year={2018}
}

@article{costa-jussa-etal-2022-no,
  title={No language left behind: Scaling human-centered machine translation},
  author={Costa-Juss{\`a}, Marta R and Cross, James and {\c{C}}elebi, Onur and Elbayad, Maha and Heafield, Kenneth and Heffernan, Kevin and Kalbassi, Elahe and Lam, Janice and Licht, Daniel and Maillard, Jean and others},
  journal={arXiv preprint arXiv:2207.04672},
  year={2022}
}

@article{de2021universal,
  title={Universal dependencies},
  author={De Marneffe, Marie-Catherine and Manning, Christopher D and Nivre, Joakim and Zeman, Daniel},
  journal={Computational linguistics},
  volume={47},
  number={2},
  pages={255--308},
  year={2021},
  publisher={MIT Press}
}

@misc{htet2025myanmarxnlibuildingdataset,
      title={Myanmar XNLI: Building a Dataset and Exploring Low-resource Approaches to Natural Language Inference with Myanmar},
      author={Aung Kyaw Htet and Mark Dras},
      year={2025},
      eprint={2504.09645},
      archivePrefix={arXiv},
      primaryClass={cs.CL},
      url={https://arxiv.org/abs/2504.09645},
}

@inproceedings{mahendra-etal-2021-indonli,
    title = "{I}ndo{NLI}: A Natural Language Inference Dataset for {I}ndonesian",
    author = "Mahendra, Rahmad  and
      Aji, Alham Fikri  and
      Louvan, Samuel  and
      Rahman, Fahrurrozi  and
      Vania, Clara",
    editor = "Moens, Marie-Francine  and
      Huang, Xuanjing  and
      Specia, Lucia  and
      Yih, Scott Wen-tau",
    booktitle = "Proceedings of the 2021 Conference on Empirical Methods in Natural Language Processing",
    month = nov,
    year = "2021",
    address = "Online and Punta Cana, Dominican Republic",
    publisher = "Association for Computational Linguistics",
    url = "https://aclanthology.org/2021.emnlp-main.821/",
    doi = "10.18653/v1/2021.emnlp-main.821",
    pages = "10511--10527",
}

@misc{nllbteam2022languageleftbehindscaling,
      title={No Language Left Behind: Scaling Human-Centered Machine Translation},
      author={NLLB Team and Marta R. Costa-juss{\`a} and James Cross and Onur {\c{C}}elebi and Maha Elbayad and Kenneth Heafield and Kevin Heffernan and Elahe Kalbassi and Janice Lam and Daniel Licht and Jean Maillard and Anna Sun and Skyler Wang and Guillaume Wenzek and Al Youngblood and Bapi Akula and Loic Barrault and Gabriel Mejia Gonzalez and Prangthip Hansanti and John Hoffman and Semarley Jarrett and Kaushik Ram Sadagopan and Dirk Rowe and Shannon Spruit and Chau Tran and Pierre Andrews and Necip Fazil Ayan and Shruti Bhosale and Sergey Edunov and Angela Fan and Cynthia Gao and Vedanuj Goswami and Francisco Guzm{\'a}n and Philipp Koehn and Alexandre Mourachko and Christophe Ropers and Safiyyah Saleem and Holger Schwenk and Jeff Wang},
      year={2022},
      eprint={2207.04672},
      archivePrefix={arXiv},
      primaryClass={cs.CL},
      url={https://arxiv.org/abs/2207.04672},
}

@inproceedings{pan-etal-2017-cross,
    title = "Cross-lingual Name Tagging and Linking for 282 Languages",
    author = "Pan, Xiaoman  and
      Zhang, Boliang  and
      May, Jonathan  and
      Nothman, Joel  and
      Knight, Kevin  and
      Ji, Heng",
    booktitle = "Proceedings of the 55th Annual Meeting of the Association for Computational Linguistics (Volume 1: Long Papers)",
    month = jul,
    year = "2017",
    address = "Vancouver, Canada",
    publisher = "Association for Computational Linguistics",
    url = "https://aclanthology.org/P17-1178/",
    doi = "10.18653/v1/P17-1178",
    pages = "1946--1958",
}

@inproceedings{ruder-etal-2021-xtreme,
    title = "{XTREME}-{R}: Towards More Challenging and Nuanced Multilingual Evaluation",
    author = "Ruder, Sebastian  and
      Constant, Noah  and
      Botha, Jan  and
      Siddhant, Aditya  and
      Firat, Orhan  and
      Fu, Jinlan  and
      Liu, Pengfei  and
      Hu, Junjie  and
      Garrette, Dan  and
      Neubig, Graham  and
      Johnson, Melvin",
    editor = "Moens, Marie-Francine  and
      Huang, Xuanjing  and
      Specia, Lucia  and
      Yih, Scott Wen-tau",
    booktitle = "Proceedings of the 2021 Conference on Empirical Methods in Natural Language Processing",
    month = nov,
    year = "2021",
    address = "Online and Punta Cana, Dominican Republic",
    publisher = "Association for Computational Linguistics",
    url = "https://aclanthology.org/2021.emnlp-main.802/",
    doi = "10.18653/v1/2021.emnlp-main.802",
    pages = "10215--10245",
}

@inbook{tiedemann2009news,
    title = "News from OPUS - A Collection of Multilingual Parallel Corpora with Tools and Interfaces",
    author = "J{\"o}rg Tiedemann",
    publisher = "University of Helsinki",
    year = "2009",
    volume = "V",
    pages = "237--248",
    booktitle = "Recent Advances in Natural Language Processing",
}

@inproceedings{zhang2020improving,
  title={Improving Massively Multilingual Neural Machine Translation and Zero-Shot Translation},
  author={Zhang, Biao and Williams, Philip and Titov, Ivan and Sennrich, Rico},
  booktitle={Proceedings of the 58th Annual Meeting of the Association for Computational Linguistics},
  pages={1628--1639},
  year={2020}
}

@inproceedings{zhang-etal-2020-improving,
  title={Improving Massively Multilingual Neural Machine Translation and Zero-Shot Translation},
  author={Zhang, Biao and Williams, Philip and Titov, Ivan and Sennrich, Rico},
  booktitle={Proceedings of the 58th Annual Meeting of the Association for Computational Linguistics},
  pages={1628--1639},
  year={2020}
}

@inproceedings{wang2019target,
  title={Target conditioned sampling: Optimizing data selection for multilingual neural machine translation},
  author={Wang, Xinyi and Neubig, Graham},
  booktitle={Proceedings of the 57th Annual Meeting of the Association for Computational Linguistics},
  pages={5823--5828},
  year={2019}
}

@article{anugraha2025r3,
  title={R3: Robust rubric-agnostic reward models},
  author={Anugraha, David and Tang, Zilu and Miranda, Lester James V and Zhao, Hanyang and Farhansyah, Mohammad Rifqi and Kuwanto, Garry and Wijaya, Derry and Winata, Genta Indra},
  journal={arXiv preprint arXiv:2505.13388},
  year={2025}
}

@inproceedings{anugraha-etal-2025-proxylm,
    title = "{P}roxy{LM}: Predicting Language Model Performance on Multilingual Tasks via Proxy Models",
    author = "Anugraha, David  and
      Winata, Genta Indra  and
      Li, Chenyue  and
      Irawan, Patrick Amadeus  and
      Lee, En-Shiun Annie",
    editor = "Chiruzzo, Luis  and
      Ritter, Alan  and
      Wang, Lu",
    booktitle = "Findings of the Association for Computational Linguistics: NAACL 2025",
    month = apr,
    year = "2025",
    address = "Albuquerque, New Mexico",
    publisher = "Association for Computational Linguistics",
    url = "https://aclanthology.org/2025.findings-naacl.106/",
    doi = "10.18653/v1/2025.findings-naacl.106",
    pages = "1981--2011",
    ISBN = "979-8-89176-195-7"
}

@inbook{Efimov_2023,
   title={The Impact of Cross-Lingual Adjustment of Contextual Word Representations on Zero-Shot Transfer},
   ISBN={9783031282416},
   ISSN={1611-3349},
   url={http://dx.doi.org/10.1007/978-3-031-28241-6_4},
   DOI={10.1007/978-3-031-28241-6_4},
   booktitle={Advances in Information Retrieval},
   publisher={Springer Nature Switzerland},
   author={Efimov, Pavel and Boytsov, Leonid and Arslanova, Elena and Braslavski, Pavel},
   year={2023},
   pages={51–67} }

@article{bert,
  author    = {Jacob Devlin and
               Ming{-}Wei Chang and
               Kenton Lee and
               Kristina Toutanova},
  title     = {{BERT:} Pre-training of Deep Bidirectional Transformers for Language
               Understanding},
  journal   = {CoRR},
  volume    = {abs/1810.04805},
  year      = {2018},
  url       = {http://arxiv.org/abs/1810.04805},
  archivePrefix = {arXiv},
  eprint    = {1810.04805},
  bibsource = {dblp computer science bibliography, https://dblp.org}
}

@misc{he2021debertav3,
      title={DeBERTaV3: Improving DeBERTa using ELECTRA-Style Pre-Training with Gradient-Disentangled Embedding Sharing}, 
      author={Pengcheng He and Jianfeng Gao and Weizhu Chen},
      year={2021},
      eprint={2111.09543},
      archivePrefix={arXiv},
      primaryClass={cs.CL}
}

@inproceedings{
marone2026mmbert,
title={mm{BERT}: A Modern Multilingual Encoder with Annealed Language Learning},
author={Marc Marone and Orion Weller and William Fleshman and Eugene Yang and Dawn Lawrie and Benjamin Van Durme},
booktitle={Forty-third International Conference on Machine Learning},
year={2026},
url={https://openreview.net/forum?id=iJDJCO4mji}
}

@article{grattafiori2024llama3herd,
  title={The Llama 3 Herd of Models},
  author={Grattafiori, Aaron and others},
  journal={arXiv preprint arXiv:2407.21783},
  year={2024}
}

\clearpage

\appendix

\section{Scope: classification tasks and choice of models}\label{appendix:scope}

We focus on classification rather than generative tasks because existing
multilingual generative benchmarks have limited language coverage. Question
Answering datasets such as TyDiQA, MLQA, and MKQA typically cover at most 11
languages \citep{clark-etal-2020-tydi, lewis-etal-2020-mlqa,
longpre-etal-2021-mkqa}, far below the 65-language coverage of the
classification benchmarks we use. A comprehensive evaluation on generative
tasks is left to future work.

For classification, encoder-only architectures remain preferred due to their
size and efficiency
\citep{dukic-snajder-2024-looking,shcharbakova-etal-2025-scale,elshabrawy-etal-2025-statement},
motivating continued investment in recent encoder-only models such as EuroBERT
\citep{boizard2025eurobert} and mmBERT \citep{marone2026mmbert}. This motivates
our main focus on XLM-R~Large. To verify that our adaptive sampling approach
generalizes beyond encoder-only models, we additionally evaluate it on Gemma~2~9B
as a decoder-only test point, following \citet{liu-niehues-2025-middle}.

\section{Related Works}

\paragraph{Cross-lingual realignment. }

Bitext-based contrastive realignment pulls together cross-lingual sentence
representations by applying a symmetric contrastive loss over in-batch negative
pairs sampled from a parallel corpus
\citep{cao2020multilingualalignmentcontextualword, wu-dredze-2020-explicit}.
This approach requires no task-specific supervision, but gains vary
substantially across languages and tasks \citep{wu-dredze-2020-explicit}.
\citet{gaschi-etal-2023-exploring} hypothesized that catastrophic forgetting is a
systematic failure mode in encoder-only models undergoing realignment;
\citet{bakos-etal-2025-alignfreeze} mitigated this with AlignFreeze, which
selectively freezes model components during training. All four works evaluated
on a limited set of predominantly high-resource or European languages.

The emergence of large-scale LRL benchmarks - driven by the Masakhane initiative
\citep{adelani-etal-2022-masakhaner2, dione-etal-2023-masakhapos,
adelani-etal-2024-irokobench} - made it newly tractable to assess realignment
across typologically diverse, low-resource languages.
\citet{nguyen-etal-2025-rethinking} aggregated those low-resource datasets with
XTREME-R \citep{hu-etal-2020-xtreme}, performing realignment in 65 languages
including 29 LRLs and evaluating on three tasks and two models. Their key
finding is that realignment is particularly impactful for LRLs — especially
those unseen during pre-training, where improvements reach up to 10 points —
while offering limited returns for high-resource languages. They also show that
a linguistically diverse language subset can match or outperform the full pool.

Our work builds directly on this line of research.
\citet{nguyen-etal-2025-rethinking} suggest that some languages might benefit
more from realignment than others, but their approach is static and binary:
decided once before training begins, with no account of how alignment difficulty
shifts across languages as training progresses. We replace this with continuous,
online per-language weights learned during realignment itself.

\paragraph{Dynamic language sampling.}

Dynamic data selection methods have been developed to better allocate training
budget based on the informativeness of samples at a given stage of learning
\citep{chen-etal-2025-scale}. At the language level, multilingual pretraining
research has studied how to adjust the mixture across languages to improve
coverage of low-resource ones
\citep{xie2023doremioptimizingdatamixtures,albalak2023efficientonlinedatamixing}.

However, dynamic sampling within the realignment phase remains underexplored.
Existing realignment methods rely on static, uniform sampling of parallel
corpora, which ignores how alignment difficulty varies across language pairs and
evolves throughout training. Our work addresses this gap by introducing an
online, loss-driven sampling strategy that adapts per-language weights during
realignment, without requiring external supervision or a held-out validation set.

\paragraph{Bilevel optimization.}

Bilevel optimization uses a hierarchical inner–outer loop structure, popularized
by MAML \citep{pmlr-v70-finn17a} for meta-learning model initialization and
extended to first-order approximations by Reptile
\citep{nichol2018firstordermetalearningalgorithms} for scalability.
\citet{ren2019learningreweightexamplesrobust} adapted this framework to dynamic
data reweighting, assigning per-example weights based on their gradient
alignment with a clean, held-out validation set.

While our inner-outer loop structure is inspired by this line of work, our
setting differs in key ways, and doesn't constitute a meta-learning setup. We do
not have access to a clean validation set with downstream labels for target
languages, so our outer loop optimizes language weights based on the realignment
loss itself, and can be seen as a form of online curriculum learning that
prioritizes hard-to-align languages, rather than a meta-learning approach that
optimizes for generalization to unseen tasks.

\section{Details about the methodology}

\subsection{Full form of the realignment loss}\label{appendix:realignment-loss}

The contrastive term $\sigma(x,y)$ used in Equation~\ref{eq:realignment} is
\begin{equation}
\sigma(x,y) = \frac{\exp\left(\frac{\text{sim}(x,y)}{T_R}\right)}{\sum_{h \in \mathcal{H},\, h \neq x} \exp\left(\frac{\text{sim}(x,h)}{T_R}\right)},
\end{equation}
where $\text{sim}$ is cosine similarity, $T_R=0.1$
\citep{wu-dredze-2020-explicit}, and $\mathcal{H}$ is the set of all source and
target representations in the batch. The symmetric pair $\log\sigma(x,y) +
\log\sigma(y,x)$ enforces that $x$ is closer to $y$ than to any other element
in $\mathcal{H}$, and vice versa.

\subsection{Gradient-based variant}\label{appendix:gradient}

For the gradient-based variant, the outer-loop loss is the weighted sum of the
accumulated per-language realignment losses over the $k$ inner steps:
\begin{equation}
    L_{\text{outer}} = \sum_{i=1}^{|\mathcal{L}_T|} w_i L_i.
\end{equation}
This loss has a closed-form maximizer that puts all the mass on the language
with the highest loss; instead we update $w$ by a single gradient-ascent step,
producing a smoother trajectory. Compared to UCB, this variant forgoes explicit
exploration but produces a smooth, differentiable update signal. The connection
between the two variants is analyzed in
Appendix~\ref{appendix:ucb-derivation}.

\subsection{Hyper-parameters}\label{appendix:hyperparams}

Hyper-parameters were fixed via a small grid search on XLM-R base. For the
UCB-based method, we set the exploration coefficient $c{=}0.1$ and the number
of inner steps $k{=}5$. For the gradient-based method, we use $k{=}10$ inner
steps and an outer-loop learning rate of $10^{-3}$, increased to $10^{-2}$ for
Gemma to compensate for the LoRA adapters' smaller gradient magnitudes.

During realignment, for XLM-R and Gemma 2, we used (respectively) a batch size of 128 and 16, a learning rate of $7.5\times10^{-6}$ and $2\times 10^{-2}$, with 16k realignment steps unless mentioned otherwise.

During fine-tuning, for XLM-R and Gemma 2, we used (respectively) a batch size of 128 and 32, the same learning rate as realignment, and during 5 epochs for each task, except NLI which is trained for 2 epochs due to its significantly larger size, following \citep{nguyen-etal-2025-rethinking}.

For LORA adapters with Gemma, we use rank $r = 8$, scaling factor $\alpha = 32$, and a LoRA dropout of $0.1$, with the adapter targeting the PEFT library's default set of attention projection modules. The same adapter is shared between the realignment and fine-tuning phases; all non-adapter parameters of the backbone are kept frozen.

\subsection{On LoRA Usage in the Decoder-Only Setting}
\label{sec:lora_usage}

\begin{table}[htbp]
\centering
\resizebox{\columnwidth}{!}{%
\begin{tabular}{l c c c}
\hline
\textbf{EXP} & \textbf{NLI Score} & \textbf{PoS Score} & \textbf{AVG} \\
\hline
Uniform realignment (w/o LoRA) & 42.47 & 31.42 & 36.94 \\
Uniform realignment (LoRA) & \textbf{61.43} & \textbf{45.68} & \textbf{53.56} \\
\hline
\end{tabular}%
}
\caption{Ablation study on the Gemma-2-2B model comparing uniform realignment with and without LoRA adapters. Notably, without LoRA, the average downstream task transfer performance reduces by nearly 17 percentage points, with the most significant detrimental effect observed on NLI tasks with a $\sim$19 percentage point reduction.}
\label{tab:lora_ablation}
\end{table}

The decision to utilize LoRA for decoder-only models in our experiments is driven by two distinct motivations. The primary constraint, as previously noted in our Limitations, is computational cost. Conducting full fine-tuning on a 9-billion parameter model over 16,000 realignment steps, followed by additional downstream fine-tuning, exceeded our available computational budget. Consequently, we followed \citet{liu-niehues-2025-middle} in adopting adapter-based tuning. 

A secondary motivation is the hypothesis that full-parameter realignment of a decoder-only model risks destabilizing its generative capabilities, potentially provoking language confusion. By employing adapter-based tuning, the core weights of the pre-trained backbone remain frozen, theoretically circumventing this instability. To empirically validate this assumption rather than relying on speculation, we conducted an ablation study comparing uniform realignment with and without LoRA using a smaller Gemma-2-2B model. 

As illustrated in Table \ref{tab:lora_ablation}, the empirical results strongly support this hypothesis. The full fine-tuning approach (without LoRA) exhibits clear signs of catastrophic forgetting, resulting in significantly lower scores. Conversely, the low-rank LoRA updates successfully preserve the model's stability, maintaining robust performance across the 2 evaluated tasks.

\subsection{Data and evaluation details}\label{appendix:data}

We follow the experimental setup of \citet{nguyen-etal-2025-rethinking}, using
OPUS-100 \citep{zhang-etal-2020-improving} and NLLB
\citep{costa-jussa-etal-2022-no} as realignment corpora and the same downstream
benchmarks. 
We distinguish four language categories: HRLs, High-Resource Languages of Joshi \citep{joshi-etal-2020-state} class 5, MRLs, Medium-Resource Languages of Joshi class 3 and 4, LRLs seen, Low-Resource Languages that were seen during pre-training of the model, and LRLs unseen, Low-Resource Languages that were not seen during pre-training. The exact language composition of each group is provided in Appendix~\ref{appendix:data}. The distinction between seen and unseen LRLs is not made for Gemma, as we are unsure of what languages were included in its pre-training data.
Table~\ref{tab:data-stats} reports, per Joshi resource class
\citep{joshi-etal-2020-state}, the number of languages evaluated, the range of
parallel training sentences used for realignment, and the size range of each
evaluation set. All 65 realignment languages are included. Rows where
\textit{Realignment n} $<$ \textit{Langs} (classes 0 and~1) contain these
eval-only languages.

\citet{nguyen-etal-2025-rethinking} originally included AmericasNLI \citep{ebrahimi-etal-2022-americasnli}, which contains languages that are not available in the parallel dataset. For this very reason, we decided not to include it in our experiments.

All experiments are run with seeds of 31, 42 and 66.

\begin{table*}
\centering
\small
\begin{adjustbox}{width=\linewidth}
\begin{tabular}{lc cc cc cc cc}
\toprule
& & \multicolumn{2}{c}{\textbf{Realignment}} & \multicolumn{2}{c}{\textbf{NER}} & \multicolumn{2}{c}{\textbf{POS}} & \multicolumn{2}{c}{\textbf{NLI}} \\
\cmidrule(lr){3-4}\cmidrule(lr){5-6}\cmidrule(lr){7-8}\cmidrule(lr){9-10}
\textbf{Joshi class} & \textbf{Langs} & \textit{n} & size range & \textit{n} & size range & \textit{n} & size range & \textit{n} & size range \\
\midrule
0 (LRL)$^*$ & 3 &  3 & 1.9M--2.8M  &  5 & 966--1{,}613  &  5 & 599--646           &  6 & 750              \\
1 (LRL) & 16 & 16 & 181k--46.4M     & 15 & 100--2{,}235  &  8 & 146--642           & 14 & 600--5{,}010     \\
2 (LRL) & 10 & 10 & 107k--32M     &  9 & 100--1{,}883  &  8 & 47--713            &  7 & 600              \\
3 (MRL) & 17 & 17 & 227k--63.6M     & 17 & 1{,}000--10{,}000 & 13 & 425--4{,}127  &  5 & 2{,}984--5{,}010 \\
4 (MRL) & 13 & 13 & 534k--1M      & 13 & 1{,}000--10{,}000 & 13 & 449--8{,}973  &  4 & 5{,}010          \\
5 (HRL) &  6 &  6 & 1M                 &  6 & 10{,}000      &  6 & 1{,}680--22{,}358 &  5 & 5{,}010          \\
\midrule
\textbf{Total} & \textbf{65} & \textbf{65} & 107{,}296--63{,}581{,}148 & \textbf{65} & --- & \textbf{53} & --- & \textbf{28} & --- \\
\textbf{Train (en)} & 1 & - & - & 1 & 20,029 & 1 & 12,570 & 1 & 392,702 \\ 
\bottomrule
\end{tabular}
\end{adjustbox}
\caption{Per-Joshi-class language coverage and evaluation set size ranges, following \citet{nguyen-etal-2025-rethinking}. \textit{Langs} counts all evaluated languages in that class; \textit{Realignment n} is the subset used for realignment (classes~0 and~1 include additional eval-only languages, hence \textit{Realignment n} $<$ \textit{Langs}). LRL = low-resource (classes 0--2), MRL = medium-resource (3--4), HRL = high-resource (5). NER: WikiANN + MasakhaNER; POS: UDPOS + MasakhaPOS; NLI: XNLI, IndoNLI, Myanmar-XNLI, and AfriXNLI. $^*$Ghomala is tentatively assigned to class~0; its Joshi class is absent from the taxonomy of \citet{joshi-etal-2020-state}.}
\label{tab:data-stats}
\end{table*}

\subsection{The link between UCB and gradient-based approaches}\label{appendix:ucb-derivation}

This section provides a more detailed discussion about the conceptual
similarities between the UCB-based and gradient-based approaches.

At each step of the outer loop, we've got a loss-per-language vector $L \in
\mathbb{R}^{|\mathcal{L}_T|}$, where $L_\ell$ is the average realignment loss
for language $\ell$ accumulated over the last $k$ inner steps. The optimization
problem that both methods are trying to solve is to find a language distribution
$p$ that maximizes the expected loss:

\begin{equation}
    \max_p \sum_{\ell \in \mathcal{L}_T} p_\ell L_\ell
\end{equation}

This is a linear optimization problem over the simplex, and its solution is to
put all the mass on the language with the highest loss. However, this would lead
to overfitting and neglecting other languages.

The gradient-based method addresses this indirectly by performing a gradient
descent step instead of directly solving the optimization problem, which leads
to a smoother update that doesn't put all the mass on a single language. 

The UCB-based update, on the other hand, is expressed as follows:

\begin{equation}
    p = \text{softmax}\left(\frac{L + c \sqrt{\frac{\ln t}{n + 1}}}{\tau}\right)
\end{equation}

If we ignore the exploration term, we get the closed-form solution of the
following optimization problem:

\begin{equation}
    \max_p \sum_{\ell \in \mathcal{L}_T} p_\ell L_\ell - \tau H(p)
\end{equation}

Where $H(p)$ is the entropy of the distribution $p$. This is a regularized
version of the original optimization problem, where the entropy term encourages
exploration and prevents putting all the mass on a single language.

Therefore, both methods can be seen as trying to solve the same underlying
optimization problem, but with different approaches to prevent overfitting and
encourage exploration. The gradient-based method does this through a smooth
update, while the UCB-based method does this through an explicit regularization
term in the optimization objective.

This also explains why the UCB-based method converges faster in the early stages
of training, as it has a more direct way to find the optimal distribution, while
the gradient-based method might be slower to find the optimal distribution due
to its indirect approach.

\section{Additional results}

\begin{table*}
\centering

\small
\adjustbox{max width=0.8\linewidth}{
\begin{tabularx}{\textwidth}{lXXXX}
\toprule
EXP & NER (F1) & NLI (Acc) & POS (Acc) & AVG \\
\midrule
\multicolumn{5}{l}{\textbf{XLM-R-Large}} \\
Fine-tuning only & 57.83 \textsubscript{$\pm$0.90} & 65.54 \textsubscript{$\pm$0.39} & 68.90 \textsubscript{$\pm$0.61} & 64.09 \textsubscript{$\pm$0.38} \\
Uniform realignment & 61.97 \textsubscript{$\pm$0.57} (+4.14) & 67.70 \textsubscript{$\pm$0.34} (+2.16) & 71.56 \textsubscript{$\pm$0.40} (+2.67) & 67.08 \textsubscript{$\pm$0.13} (+2.99) \\
Most-URIEL & 62.00 \textsubscript{$\pm$0.56} (+4.17) & 67.98 \textsubscript{$\pm$0.32} (+2.44) & 70.93 \textsubscript{$\pm$0.35} (+2.03) & 66.97 \textsubscript{$\pm$0.37} (+2.88) \\
Gradient-based & \textbf{62.25 \textsubscript{$\pm$1.10} (+4.42)} & 68.65 \textsubscript{$\pm$0.29} (+3.11) & \textbf{72.33 \textsubscript{$\pm$0.34} (+3.44)} & \textbf{67.75 \textsubscript{$\pm$0.31} (+3.66)} \\
UCB-based & 61.93 \textsubscript{$\pm$0.79} (+4.09) & \textbf{68.94 \textsubscript{$\pm$0.02} (+3.40)} & 72.33 \textsubscript{$\pm$0.42} (+3.44) & 67.73 \textsubscript{$\pm$0.13} (+3.64) \\
\midrule
\multicolumn{5}{l}{\textbf{Gemma-2-9b}} \\
Fine-tuning only & 36.09 \textsubscript{$\pm$1.18} & 63.30 \textsubscript{$\pm$3.67} & 49.09 \textsubscript{$\pm$0.94} & 49.49 \textsubscript{$\pm$1.42} \\
Uniform realignment & 36.51 \textsubscript{$\pm$1.35} (+0.42) & 69.97 \textsubscript{$\pm$1.08} (+6.67) & 48.22 \textsubscript{$\pm$4.70} (-0.87) & 51.57 \textsubscript{$\pm$2.24} (+2.07) \\
Most-URIEL & \textbf{37.60 \textsubscript{$\pm$1.17} (+1.51)} & \textbf{70.11 \textsubscript{$\pm$1.01}} (+6.81) & \textbf{51.83 \textsubscript{$\pm$0.85} (+2.74)} & \textbf{53.18 \textsubscript{$\pm$0.28} (+3.69)} \\
Gradient-based & 34.54 \textsubscript{$\pm$3.94} (-1.55) & 69.31 \textsubscript{$\pm$0.81} (+6.01) & 50.15 \textsubscript{$\pm$1.41} (+1.06) & 51.33 \textsubscript{$\pm$1.45} (+1.84) \\
UCB-based & 37.06 \textsubscript{$\pm$0.30} (+0.97) & 69.36 \textsubscript{$\pm$0.36} (+6.07) & 50.06 \textsubscript{$\pm$5.97} (+0.97) & 52.16 \textsubscript{$\pm$1.87} (+2.67) \\
\midrule
\multicolumn{5}{l}{\textbf{mBERT}} \\
Fine-tuning only & 52.85 \textsubscript{$\pm$0.25} & 62.94 \textsubscript{$\pm$0.08} & 53.03 \textsubscript{$\pm$0.04} & 56.27 \textsubscript{$\pm$0.09} \\
Uniform realignment & \textbf{56.33 \textsubscript{$\pm$0.23} (+3.48)} & 68.82 \textsubscript{$\pm$0.06} (+5.88) & 58.40 \textsubscript{$\pm$0.07} (+5.36) & \textbf{61.18 \textsubscript{$\pm$0.08} (+4.91)} \\
Most-URIEL & 55.76 \textsubscript{$\pm$0.17} (+2.91) & 67.92 \textsubscript{$\pm$0.35} (+4.99) & 58.43 \textsubscript{$\pm$0.21} (+5.40) & 60.70 \textsubscript{$\pm$0.15} (+4.43) \\
Gradient-based & 55.39 \textsubscript{$\pm$0.65} (+2.54) & \textbf{69.05 \textsubscript{$\pm$0.10} (+6.11)} & 58.72 \textsubscript{$\pm$0.37} (+5.69) & 61.05 \textsubscript{$\pm$0.25} (+4.78) \\
UCB-based & 56.05 \textsubscript{$\pm$0.47} (+3.20) & 68.71 \textsubscript{$\pm$0.06} (+5.78) & \textbf{58.76 \textsubscript{$\pm$0.34} (+5.72)} & 61.17 \textsubscript{$\pm$0.25} (+4.90) \\
\midrule
\multicolumn{5}{l}{\textbf{mmBERT}} \\
Fine-tuning only & 45.32 \textsubscript{$\pm$1.97} & 58.64 \textsubscript{$\pm$1.99} & 60.21 \textsubscript{$\pm$0.44} & 54.72 \textsubscript{$\pm$1.30} \\
Uniform realignment & 51.16 \textsubscript{$\pm$1.20} (+5.83) & \textbf{68.99 \textsubscript{$\pm$0.30} (+10.34)} & 63.48 \textsubscript{$\pm$1.18} (+3.27) & 61.21 \textsubscript{$\pm$0.78} (+6.48) \\
Most-URIEL & 51.81 \textsubscript{$\pm$0.43} (+6.48) & 68.09 \textsubscript{$\pm$0.74} (+9.45) & 63.14 \textsubscript{$\pm$0.25} (+2.94) & 61.01 \textsubscript{$\pm$0.46} (+6.29) \\
Gradient-based & \textbf{53.59 \textsubscript{$\pm$2.56} (+8.27)} & 68.64 \textsubscript{$\pm$0.16} (+10.00) & 63.50 \textsubscript{$\pm$0.67} (+3.29) & \textbf{61.91 \textsubscript{$\pm$0.70} (+7.19)} \\
UCB-based & 52.48 \textsubscript{$\pm$1.71} (+7.15) & 68.26 \textsubscript{$\pm$0.82} (+9.62) & \textbf{63.69 \textsubscript{$\pm$0.29} (+3.48)} & 61.48 \textsubscript{$\pm$0.76} (+6.75) \\
\midrule
\multicolumn{5}{l}{\textbf{mDeBERTa-v3}} \\
Fine-tuning only & 66.88 \textsubscript{$\pm$0.22} & \textbf{60.87 \textsubscript{$\pm$0.77}} & 71.35 \textsubscript{$\pm$0.26} & 66.36 \textsubscript{$\pm$0.30} \\
Uniform realignment & 68.33 \textsubscript{$\pm$0.05} (+1.45) & 59.72 \textsubscript{$\pm$0.88} (-1.15) & \textbf{72.44 \textsubscript{$\pm$0.47} (+1.09)} & 66.83 \textsubscript{$\pm$0.35} (+0.46) \\
Most-URIEL & 68.09 \textsubscript{$\pm$0.16} (+1.21) & 58.83 \textsubscript{$\pm$0.45} (-2.04) & 71.34 \textsubscript{$\pm$0.56} (-0.00) & 66.09 \textsubscript{$\pm$0.03} (-0.28) \\
Gradient-based & \textbf{68.79 \textsubscript{$\pm$0.16} (+1.91)} & 60.04 \textsubscript{$\pm$1.26} (-0.83) & 71.72 \textsubscript{$\pm$0.88} (+0.37) & 66.85 \textsubscript{$\pm$0.25} (+0.49) \\
UCB-based & 68.53 \textsubscript{$\pm$0.30} (+1.65) & 60.30 \textsubscript{$\pm$0.87} (-0.57) & 72.05 \textsubscript{$\pm$0.40} (+0.70) & \textbf{66.96 \textsubscript{$\pm$0.21} (+0.59)} \\
\midrule
\multicolumn{5}{l}{\textbf{Llama 3.1 8b}} \\
Fine-tuning only (LoRA) & 40.36 \textsubscript{$\pm$2.72} & 56.55 \textsubscript{$\pm$2.63} & 29.98 \textsubscript{$\pm$0.62} & 42.30 \textsubscript{$\pm$0.33} \\
Uniform realignment (LoRA) & 41.50 \textsubscript{$\pm$6.10} (+1.14) & 62.01 \textsubscript{$\pm$1.09} (+5.45) & 31.79 \textsubscript{$\pm$2.01} (+1.81) & 45.10 \textsubscript{$\pm$2.92} (+2.80) \\
Most-URIEL & 42.58 \textsubscript{$\pm$2.53} (+2.22) & 62.01 \textsubscript{$\pm$0.83} (+5.46) & 31.86 \textsubscript{$\pm$1.92} (+1.88) & 45.48 \textsubscript{$\pm$1.52} (+3.19) \\
Gradient-based (LoRA) & \textbf{43.24 \textsubscript{$\pm$3.10} (+2.88)} & 61.50 \textsubscript{$\pm$1.76} (+4.94) & \textbf{32.67 \textsubscript{$\pm$0.72} (+2.69)} & \textbf{45.80 \textsubscript{$\pm$1.36} (+3.50)} \\
UCB-based (LoRA) & 41.26 \textsubscript{$\pm$5.41} (+0.91) & \textbf{62.28 \textsubscript{$\pm$0.89} (+5.73)} & 32.04 \textsubscript{$\pm$1.02} (+2.06) & 45.20 \textsubscript{$\pm$2.41} (+2.90) \\
\bottomrule
\end{tabularx}
}
\caption{Average performance across three random seeds for each task and across all three tasks, comparing three baselines with two dynamic sampling strategies. All realignment experiments use 16{,}000 realignment steps.}
\label{tab:res_all}
\end{table*}

\subsection{Dynamic Languages Weights Analysis}
\label{sec:analysis}

\begin{figure*}
    \centering
    \begin{subfigure}[t]{0.48\linewidth}
        \includegraphics[width=\linewidth]{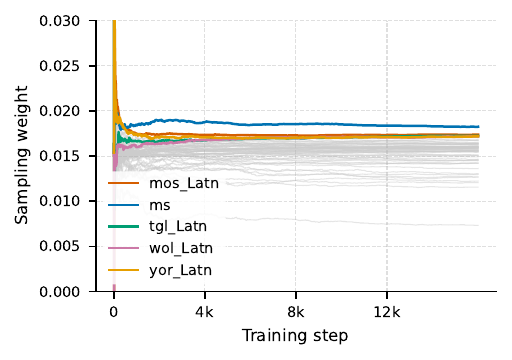}
        \caption{UCB weights, seed 31}
    \end{subfigure}
    \hfill
    \begin{subfigure}[t]{0.48\linewidth}
        \includegraphics[width=\linewidth]{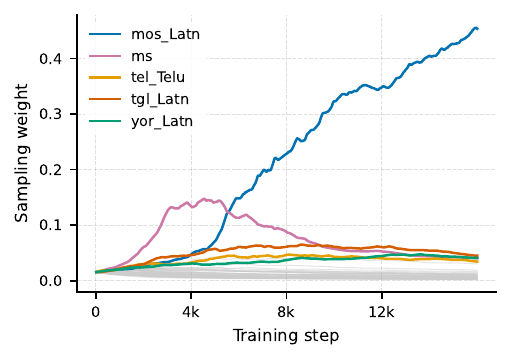}
        \caption{Gradient weights, seed 31}
    \end{subfigure}

    \begin{subfigure}[t]{0.48\linewidth}
        \includegraphics[width=\linewidth]{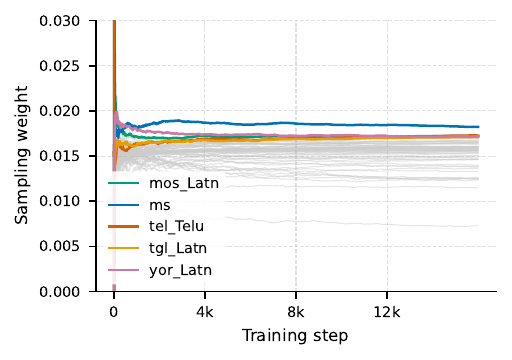}
        \caption{UCB weights, seed 42}
    \end{subfigure}
    \hfill
    \begin{subfigure}[t]{0.48\linewidth}
        \includegraphics[width=\linewidth]{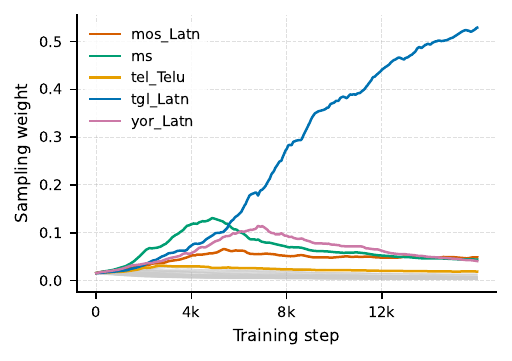}
        \caption{Gradient weights, seed 42}
    \end{subfigure}

    \begin{subfigure}[t]{0.48\linewidth}
        \includegraphics[width=\linewidth]{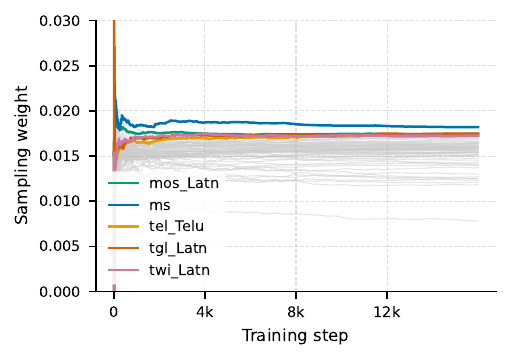}
        \caption{UCB weights, seed 66}
    \end{subfigure}
    \hfill
    \begin{subfigure}[t]{0.48\linewidth}
        \includegraphics[width=\linewidth]{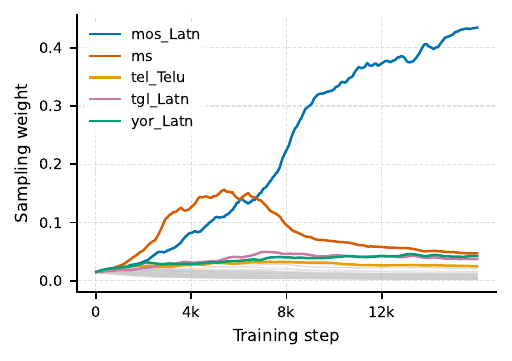}
        \caption{Gradient weights, seed 66}
    \end{subfigure}

    \caption{Learned weights for Gemma 2 9B across seeds for UCB (left) and gradient-based (right) selection strategies.}
    \label{fig:appendix-weights-comparison-gemma}
\end{figure*}

To better understand the dynamics of our proposed strategies,
Figures~\ref{fig:appendix-weights-comparison}
and~\ref{fig:appendix-weights-comparison-gemma} plot the evolution of language
weights across training for XLM-R and Gemma 2 9B respectively. Both UCB and
gradient-based methods oversample roughly the same top-5 set of languages per
model, despite their different designs, and this ordering is consistent across
seeds. For Gemma, the oversampled
languages are all LRLs except Malay (ms). For XLM-R Large, they include
mid-resource languages like Hebrew (he), low-resource ones like Tswana
(tsn\_Latn) and Tamil (ta), and typologically distant ones like Malay (ms) and
Hindi (hi), though Hindi is Indo-European like English. Since this
oversampling yields better downstream results, this supports our hypothesis
that realignment benefits from oversampling LRLs.

The two methods differ in how the weights evolve. For XLM-R, UCB quickly
converges to a stable distribution with slower subsequent dynamics (e.g., the
gradual increase of Hindi), while the gradient-based method appears
monotonically divergent without stabilizing, though some non-monotonic
evolution is visible for Malay, especially with Gemma 2. This reflects a key
limitation: gradient-based has no explicit exploration mechanism and may
over-exploit early winners, whereas UCB explores by design.

More importantly, the gradient-based method is slower in early stages but
suffers from strong momentum later. For XLM-R, Hebrew reaches a sampling
probability above 0.04 in under 1k steps with UCB, but takes roughly 10$\times$
longer with gradient-based; it then keeps rising with gradient-based while
slowly diminishing with UCB. This likely explains why UCB performs better on
longer runs while gradient-based degrades at 32k steps
(Figure~\ref{fig:scaling-graph}). The slower early learning of the
gradient-based method may stem from it being a single gradient step, whereas
UCB's update closely resembles the closed-form solution to the regularized
problem of maximizing the realignment loss
(Appendix~\ref{appendix:ucb-derivation}).

\subsection{Realignment Step Scaling}
\begin{figure}[!th]
    \centering
    \includegraphics[width=\columnwidth]{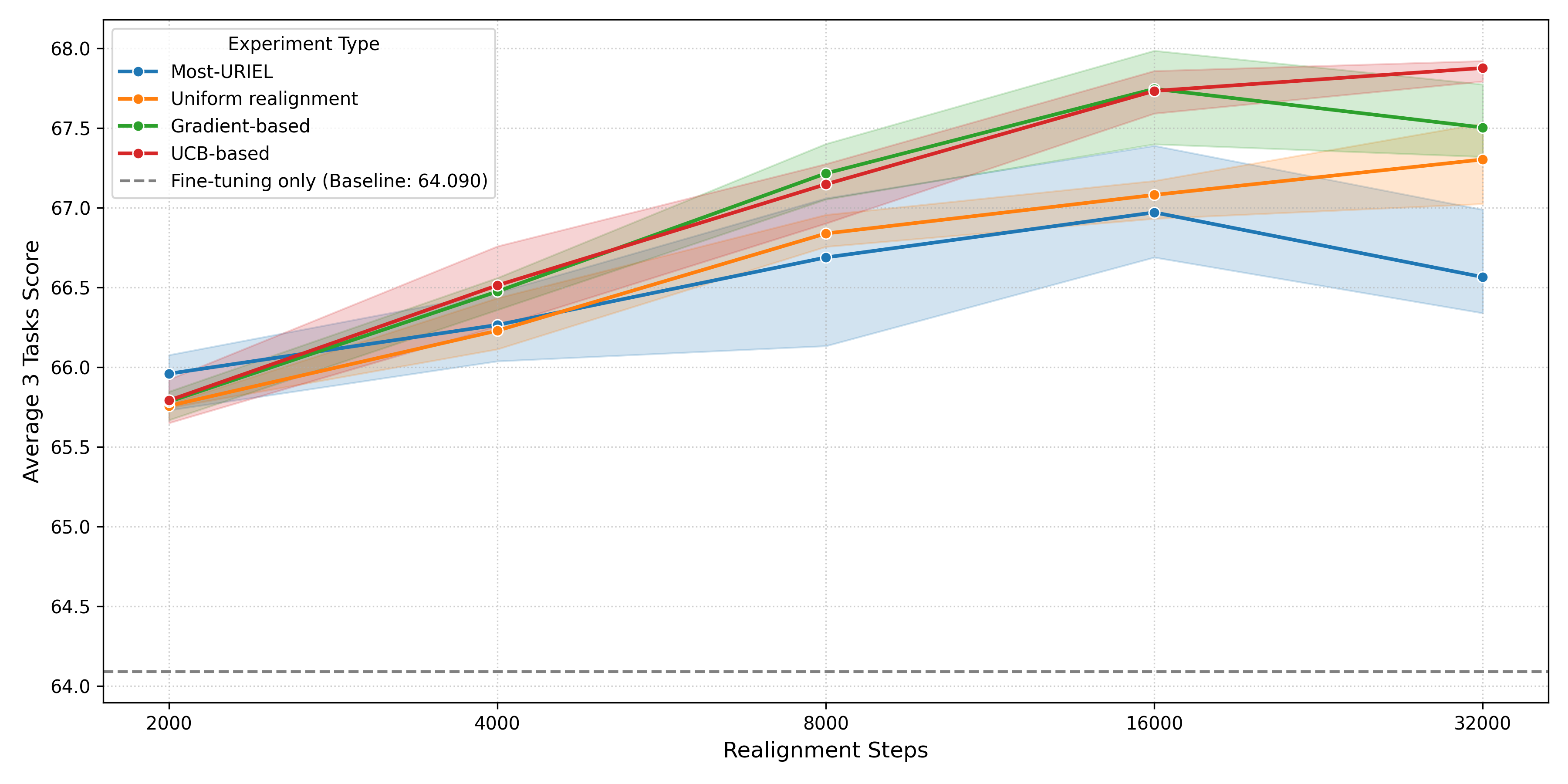}
    \caption{Scaling of average cross-lingual transfer performance for XLM-R}
    \label{fig:scaling-graph}
\end{figure}

\begin{figure*}
    \centering
    \begin{subfigure}[t]{0.48\linewidth}
        \includegraphics[width=\linewidth]{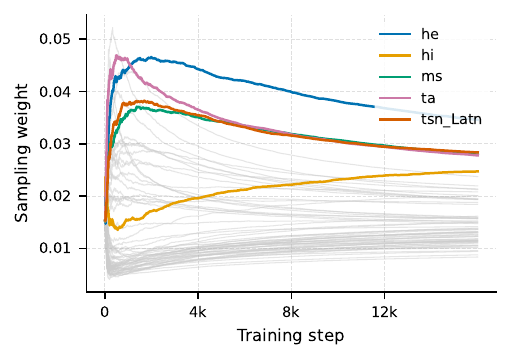}
        \caption{UCB weights, seed 31}
    \end{subfigure}
    \hfill
    \begin{subfigure}[t]{0.48\linewidth}
        \includegraphics[width=\linewidth]{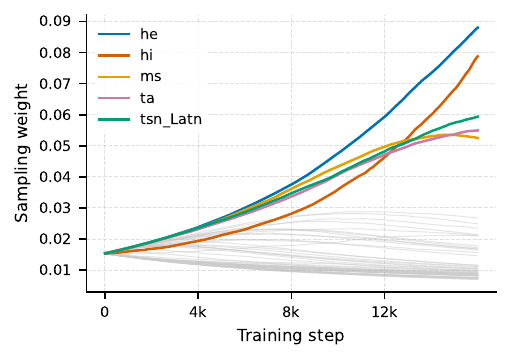}
        \caption{Gradient weights, seed 31}
    \end{subfigure}

    \begin{subfigure}[t]{0.48\linewidth}
        \includegraphics[width=\linewidth]{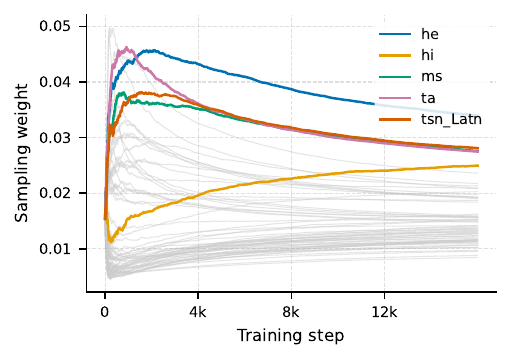}
        \caption{UCB weights, seed 42}
    \end{subfigure}
    \hfill
    \begin{subfigure}[t]{0.48\linewidth}
        \includegraphics[width=\linewidth]{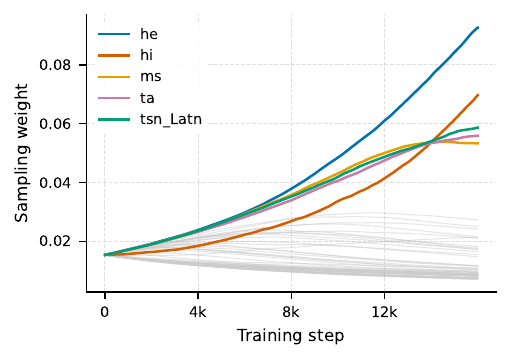}
        \caption{Gradient weights, seed 42}
    \end{subfigure}

    \begin{subfigure}[t]{0.48\linewidth}
        \includegraphics[width=\linewidth]{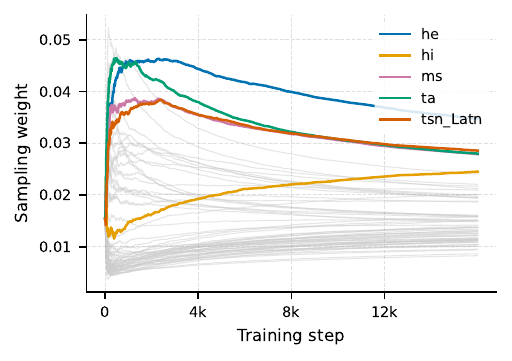}
        \caption{UCB weights, seed 66}
    \end{subfigure}
    \hfill
    \begin{subfigure}[t]{0.48\linewidth}
        \includegraphics[width=\linewidth]{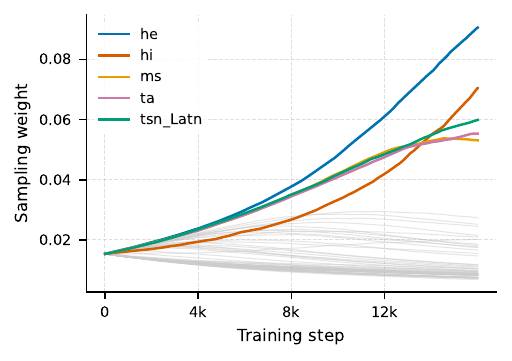}
        \caption{Gradient weights, seed 66}
    \end{subfigure}

    \caption{Learned weights for XLM-R Large across seeds for UCB (left) and gradient-based (right) selection strategies.}
    \label{fig:appendix-weights-comparison}
\end{figure*}

Most-URIEL achieves stronger gains than the other strategies at low compute
budgets (2k steps), but the gap widens in favor of the dynamic strategies as
steps increase (Figure~\ref{fig:scaling-graph}). At $2{,}000$ steps, Most-URIEL
leads with an average score of roughly $66.0$ — approximately $0.2$ points ahead of
the dynamic methods. This reflects the data-efficiency of typologically-informed
static selection: with scarce compute, every step targets a pre-curated diverse
language subset, consistent with the finding that linguistic diversity is a key
driver of language subset quality for realignment
\citep{nguyen-etal-2025-rethinking}. Dynamic methods, starting from a
near-uniform distribution, seem to require a warm-up period for their weights to
concentrate toward hard-to-align languages, incurring an early exploration cost.
As training continues, however, the static selection's advantage erodes. At
16{,}000 steps, UCB and gradient-based sampling both reach an average of
67.73--67.75 (Table~\ref{tab:res_overview}), surpassing Most-URIEL's 66.97 by
nearly 0.8 points. This crossover directly reflects the rigidity of static
distributions: since alignment difficulty shifts throughout training (see
Figure~\ref{fig:appendix-weights-comparison}), a distribution fixed before training
begins becomes progressively misaligned with the model's evolving
needs~\citep{albalak2023efficientonlinedatamixing}. Most importantly, this shows
that dynamic sampling scales better with training time, as it can adapt to the
model's changing alignment needs, while static sampling cannot.

Gradient-based degrades at 32k steps, while UCB remains stable — pointing to
UCB's greater robustness to training duration as a key advantage. Concretely,
the gradient-based method drops from $67.75$ at $16{,}000$ steps to approximately
$67.50$ at $32{,}000$, while UCB continues to improve to roughly $67.9$. We attribute
this to hyperparameter sensitivity: all hyperparameters were tuned via grid
search at $16{,}000$ steps, and gradient-based seems to have overfit on this
setting. This brittleness echoes the behaviour observed on Gemma
(Section~\ref{results}), where gradient-based also underperformed — in both
cases, performance degrades when conditions diverge from the tuning regime. UCB
is less exposed to this issue: its update rule has a principled closed-form
character (Appendix~\ref{appendix:ucb-derivation}) and its single exploration
coefficient $c$ proves less sensitive to the choice of training length.

Nevertheless, we find 16{,}000 realignment steps to provide the best trade-off
between computational cost and performance, as doubling the training time to
32{,}000 steps yields only an additional gain of roughly 0.4 points. These
findings provide practical guidance for selecting realignment strategies under
different resource constraints.

\subsection{Language-level win rate analysis}\label{appendix:winloss}

\begin{figure*}[!th]
    \centering
    \includegraphics[width=\linewidth]{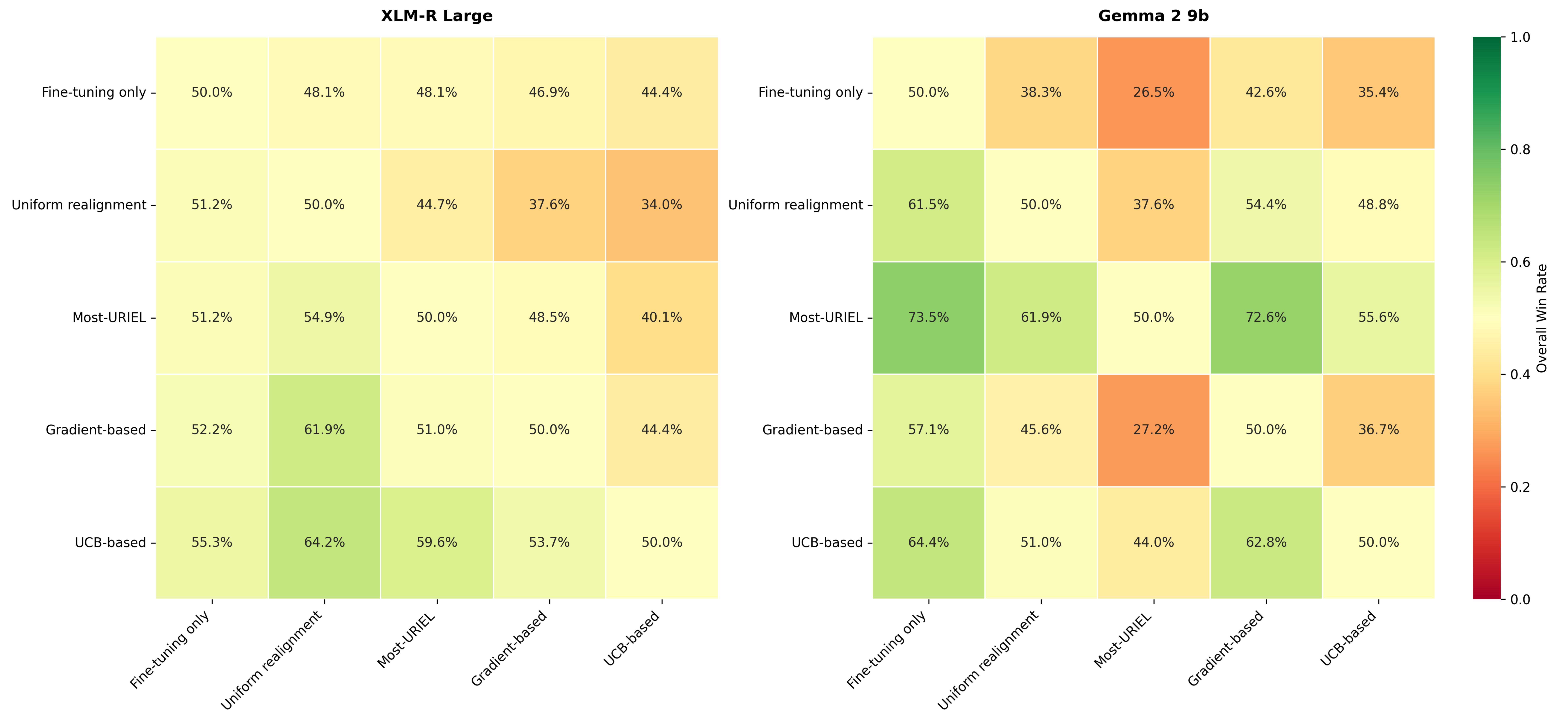}
    \caption{Win Rate, N=441 (all tasks, all languages, all seeds). The value of a cell indicates how many times the method in the row outperformed the method in the column.}
    \label{fig:win-loss-figure}
\end{figure*}

\begin{figure*}[!th]
    \centering
    \includegraphics[width=\linewidth]{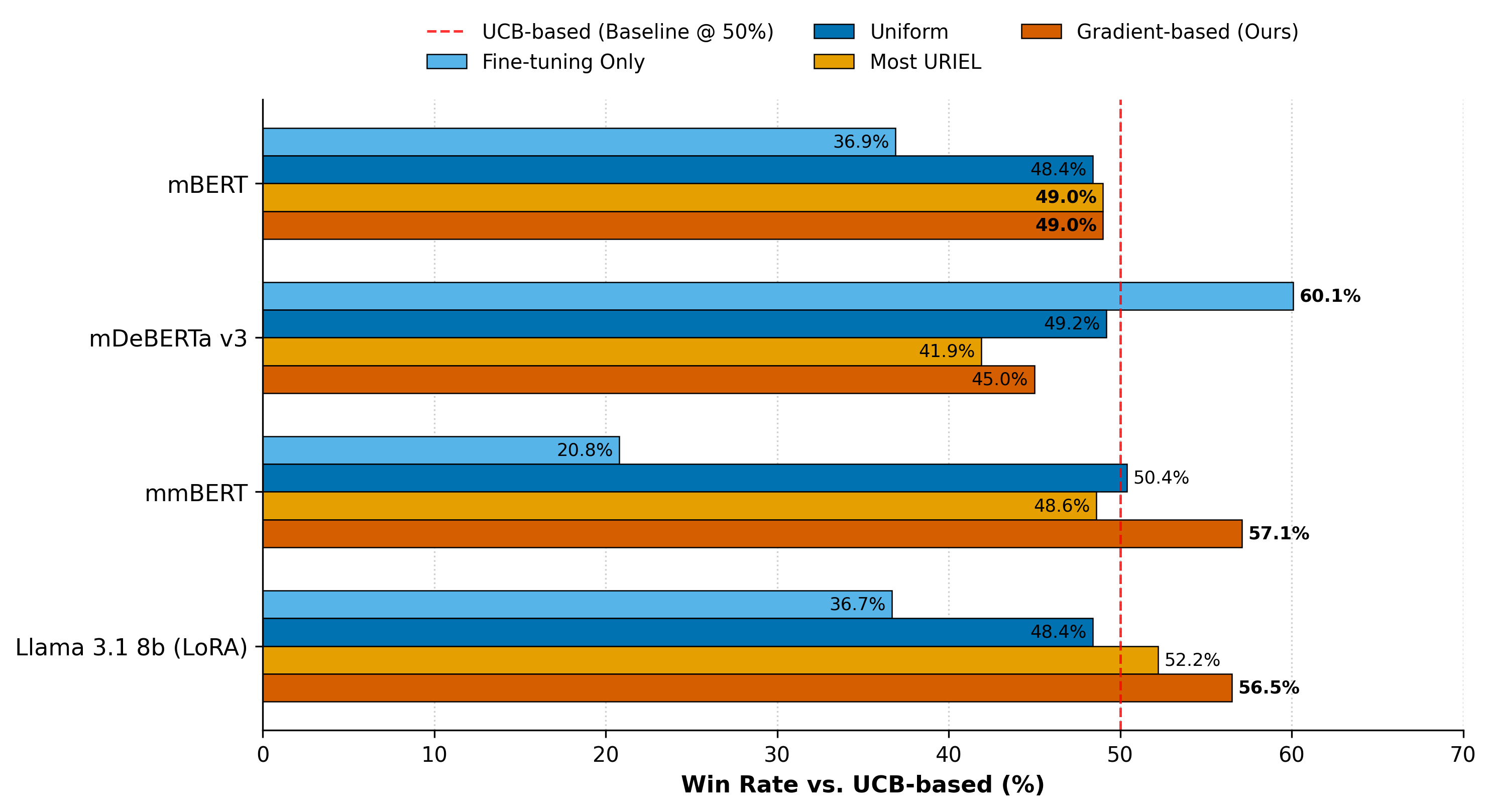}
    \caption{Win Rate comparing other experiments to our proposed UCB-based method for extra 3 encoder-only models (mBERT, mDeBERTa-v3 and mmBERT) and 1 decoder-only model (Llama 3.1 8b) on the same tasks, languages and seeds. Bold indicates the the best win-rate of each model. Uniform realignment cannot surpass UCB-based strategy in distributing realignment gain across more languages in most of the cases.}
    \label{fig:extra-win-loss}
\end{figure*}

To complement the average performance analysis, we also examine the win rates
of the different methods. This win rate is computed, given two methods A and
B, as the percentage of runs, represented by a unique task, language, and
seed, for which method A outperforms method B. This analysis allows us to
understand how consistent the performance gains are across languages, and
whether the improvements are driven by a few languages or are more widespread.

It is in the win-rate analysis that the positive impact of dynamic sampling
becomes most apparent (Figure~\ref{fig:win-loss-figure}). Both dynamic
sampling methods further improve cross-lingual transfer gains over the static
baselines (uniform and Most-URIEL) while also enhancing language-wise
performance. In particular, both dynamic sampling methods outperform uniform
sampling by at least $61.9\%$. This is coherent with resource-level breakdown
results, where we observed that dynamic sampling spreads the gains across more
languages.

In the case of Gemma, all realignment methods win versus fine-tuning, and
Most-URIEL is a surprisingly strong baseline, which mirrors the overall
results. However, the win-rates show that UCB-based dynamic sampling still
wins over uniform sampling, and is second only to Most-URIEL. Moreover, the
gradient-based method falls behind. This suggests that a set of
hyperparameters that works for one model will probably not be optimal for
another one, since we tuned ours with XLM-R. As we already saw earlier, the
UCB-based method generalizes better to other architectures than the
gradient-based one.

For Gemma, realignment itself is highly effective, and while UCB is the most
reliable dynamic approach, Most-URIEL sets a high bar that UCB doesn't clearly
surpass. When working with a new model architecture, using a subset of
languages might be a good starting point before trying to implement a dynamic
sampling strategy.

We additionally present a win-rate comparison of our proposed UCB-based method across the four supplementary models (three encoder-only and one decoder-only). As illustrated in Figure~\ref{fig:extra-win-loss}, the UCB-based strategy establishes a formidable baseline, outperforming uniform realignment on the majority of models - with uniform sampling edging it out by a marginal 0.4\% on mmBERT. The approach to most frequently surpass it (on 2 of the 4 models) is our alternative gradient-based method. Although the gradient-based strategy yields limited gains over the uniform sampling baseline on Gemma-2-9B, it achieves superior performance on both mmBERT and Llama 3.1 8B.

\section{Computational resources}

All experiments with XLM-R Large and Gemma 2 9B were run on a single H100 80GB
GPU. The total compute used for all experiments is estimated to be around 1,000
GPU hours. 

A typical realignment run with XLM-R for 16k steps takes around 4.5 hours, while
a typical run with Gemma 2 9B takes around 7 hours (not including fine-tuning).

\section{Licenses for artifacts used}
\label{sec:licenses}
Below is a list of the datasets under study:
\begin{itemize}
    \item The AfriXNLI dataset~\citep{adelani-etal-2024-irokobench} has the Apache 2.0 license.
    \item The IndoNLI dataset~\citep{mahendra-etal-2021-indonli} has the CC-BY-SA 4.0 license.
    \item The Myanmar-XNLI dataset~\citep{htet2025myanmarxnlibuildingdataset} has the Apache 2.0 license.
    \item The MasakhaPOS dataset~\citep{dione-etal-2023-masakhapos} has the MIT license.
    \item The MasakhaNER 2.0 dataset~\citep{adelani-etal-2022-masakhaner2} has the AFL 3.0 license.
    \item The OPUS-100 dataset~\citep{zhang2020improving} has no explicit license; it is a filtered subset of OPUS~\citep{tiedemann2009news}, which aggregates translation corpora that is generally considered redistributable.  
    \item The NLLB dataset~\citep{nllbteam2022languageleftbehindscaling} has the ODC-By license.
    \item The XTREME-R benchmark suite~\citep{ruder-etal-2021-xtreme} does not have a unified license; it aggregates multiple datasets, each with its own license or terms of use, here are the ones we use:\\
    \begin{itemize}
        \item The XNLI corpus~\citep{conneau2018xnli} has the CC BY-NC 4.0 license.
        \item The UDPOS dataset~\citep{de2021universal} has the CC0-1.0 license.
        \item The WikiANN dataset~\citep{pan-etal-2017-cross} has the Apache 2.0 license.
    \end{itemize}
\end{itemize}

Below is a list of the other artifacts under study:
\begin{itemize}
    \item The code for realignment comes from \citet{nguyen-etal-2025-rethinking} and has the MIT license.
    \item The weights of XLM-R Base ~\citep{conneau-etal-2020-unsupervised} have the MIT license.
    \item The weights of Gemma 2 9B ~\citep{gemmateam2024gemma2improvingopen}
    have a dedicated Gemma License, which allows for research use and
    redistribution under certain conditions.
\end{itemize}
All artifacts were thus used in accordance with their open-source or non-commercial licenses.

\section{Use of AI}

For the writing of this paper, AI was used for the following purposes:
reformulate some text, autocomplete code, help with technical issues for Python
code and LaTeX, and occasionally help with brainstorming and for generating code
that would produce the figures. 

All the ideas of the paper were proposed by the authors. All the text was
originally written and revised by the authors, with some reformulation help from
AI. All generated code was revised and edited by the authors. No figure was
directly generated by AI.

\end{document}